\documentclass{article}
\usepackage{arxiv}

\makeatletter
\renewcommand{\today}{}
\renewcommand{\@date}{}
\makeatother

\usepackage[utf8]{inputenc}
\usepackage{amsmath}
\usepackage{amsfonts}
\usepackage{amssymb}
\usepackage{graphicx}
\usepackage{booktabs}
\usepackage{multirow}
\usepackage{algorithm}
\usepackage{algorithmic}
\usepackage{xcolor}
\usepackage{tikz}
\usepackage{pgfplots}
\usepackage{subcaption}
\usepackage[hidelinks]{hyperref}
\usepackage{geometry}
\usepackage{listings}
\usepackage{float}
\usepackage{color,soul}
\usepackage[breakable]{tcolorbox}
\usepackage{array}
\usepackage{enumitem}

\pgfplotsset{compat=1.17}

\title{\textbf{Mapping Patient-Perceived Physician Traits from Nationwide Online Reviews with LLMs}}

\author{
\textbf{Junjie Luo}$^{1}$,
\textbf{Rui Han}$^{3,4}$,
\textbf{Arshana Welivita}$^{2}$,
\textbf{Zeleikun Di}$^{3}$,\\
\textbf{Jingfu Wu}$^{3}$,
\textbf{Xuzhe Zhi}$^{2}$,
\textbf{Tianli Xu}$^{3}$,
\textbf{Ritu Agarwal}$^{3,4}$,
\textbf{Gordon Gao}$^{3,4,*}$ \\
\\
$^1$Johns Hopkins School of Medicine, Baltimore, MD, USA. \\
$^2$Johns Hopkins University, Baltimore, MD, USA. \\
$^3$Carey Business School, Johns Hopkins University, Baltimore, MD, USA. \\
$^4$Center for Digital Health Artificial Intelligence (CDHAI), Baltimore, MD, USA. \\
\\
*Correspondence: gordon.gao@jhu.edu
}

\begin{document}

\maketitle


\begin{abstract}
%
%
Understanding how patients perceive their physicians is essential to improving trust, communication, and satisfaction.
%
Patients increasingly consult large language models (LLMs) to summarize physician reviews and shape provider choices, yet the national landscape of patient-perceived physician traits remains poorly characterized.
%
We present an LLM-based pipeline that extracts ten patient-perceived physician trait scores from review text: five Big-Five-style and five patient-oriented dimensions.
%
From one million U.S. physicians, we analyze 4.1 million reviews of 226,999 physicians.
%
We validate the pipeline through multi-model comparison and human expert benchmarking.
%
LLM and human-rater trait scores from reviews are consistent.
%
Trait scores correlate strongly with review rating scores yet retain substantial independent variance.
%
Two national-scale patterns emerge: male physicians receive higher trait scores across all traits, with the largest gap in clinical competence; specialty differences are driven by encounter context, with surgical specialties leading interpersonal qualities and psychiatry lowest.
%
Cluster analysis identifies four physician archetypes, from ``Uniform High'' (33.8\%, high across traits) to ``Uniform Low'' (22.6\%, low across traits).
%
This map of LLM-derived physician traits exposes how LLMs read the U.S. clinical workforce.
%
Pending clinical validation, it opens future research on fairness, bias, and how LLM-mediated provider search shapes patient choice.
\end{abstract}


\section{Introduction}
%
%

Interpersonal and professional qualities of physicians profoundly shape patient trust, communication, adherence, and health outcomes~\cite{lerch2024model, wu2022relationship}.
%
Yet current measurement tools, such as standardized surveys and aggregate star ratings, capture a narrow view of the physician-patient relationship from the patient's perspective.
%
Meanwhile, millions of online physician reviews provide a patient-generated record of real-world experiences at national scale, encoding rich interpersonal and professional signal about empathy, competence, and attentiveness that surveys do not capture~\cite{mcgrath2018validity, wang2022recency, schlesinger2015taking, hanauer2014public}.
%
Yet despite their clinical importance, the national landscape of these patient-perceived traits has never been mapped.

%
Two categories capture patient-perceived physician traits: psychological traits in the Big-Five tradition, and care-specific judgments such as communication, competence, attentiveness, and trustworthiness.
%
Despite their established clinical relevance~\cite{john2008paradigm, atherton2014personality}, these traits have not been characterized at national scale from review text.
%
Existing large-scale review studies operated on aggregate ratings or coarse text axes such as sentiment and topic modeling~\cite{gao2012landscape, daskivich2018differences}, while richer interpersonal coding has remained manual and small-sample~\cite{dunivin2020gender}.
%
The national patient-perceived trait landscape thus remains uncharacterized.
%

%
Recently, patients increasingly consult large language models (LLMs) such as ChatGPT and Gemini to search for physicians, summarize patient reviews, and weigh tradeoffs across providers~\cite{littrell2025patients,tebra2026choose,landi2026chatgpt,ayo2024characterizing,mendel2025laypeople}.
%
When an LLM reads thousands of reviews and renders an opinion about a physician's empathy or competence, that opinion increasingly mediates patient choice.
%
Whether LLM-derived physician traits are consistent with patient-perceived traits, and which LLM models produce extractions closest to human perception, has not been measured.
%
Auditing this consistency is therefore a prerequisite for credible national characterization of patient-perceived traits, with fairness implications for LLM-mediated provider search.
%

%
We therefore ask what the national landscape of patient-perceived physician traits looks like across the U.S. physician workforce: their distributions, demographic correlates, and latent archetypes.
%
Because manual annotation at this scale is infeasible, we use LLMs to extract these traits and ask whether the extractions are consistent with patient-perceived traits, and which LLM comes closest to human perception~\cite{minaee2024large}.

%
Here, we present and validate an LLM-based pipeline that extracts ten patient-perceived traits from millions of online physician reviews: five Big-Five-style interpersonal dimensions and five patient-oriented subjective judgments.
%
We assess the consistency between LLM-derived and patient-perceived traits through multi-model comparison, an LLM-as-a-Judge framework with a cross-family judge, and Human-as-a-Judge labels on random and hard-case panels.
%
We then apply the selected pipeline to 226{,}999 U.S. physicians~\cite{zheng2023judging}.
%
We characterize national trait distributions, explore correlations between Big-Five-style and patient-oriented trait scores, examine differences by gender and specialty, and identify latent physician archetypes through cluster analysis.

%
These results may inform future research on how physician interpersonal style shapes patient experience and on fairness in LLM-mediated patient evaluation.
%
Clinical validity of these trait profiles is a separate question requiring independent study.
%
The results below cover dataset characteristics, the consistency assessment of LLM-derived traits against human perception, national trait distributions, demographic and specialty patterns, and physician archetypes.

\section{Results}
\begin{figure}[htbp]
\centering
\includegraphics[width=\textwidth]{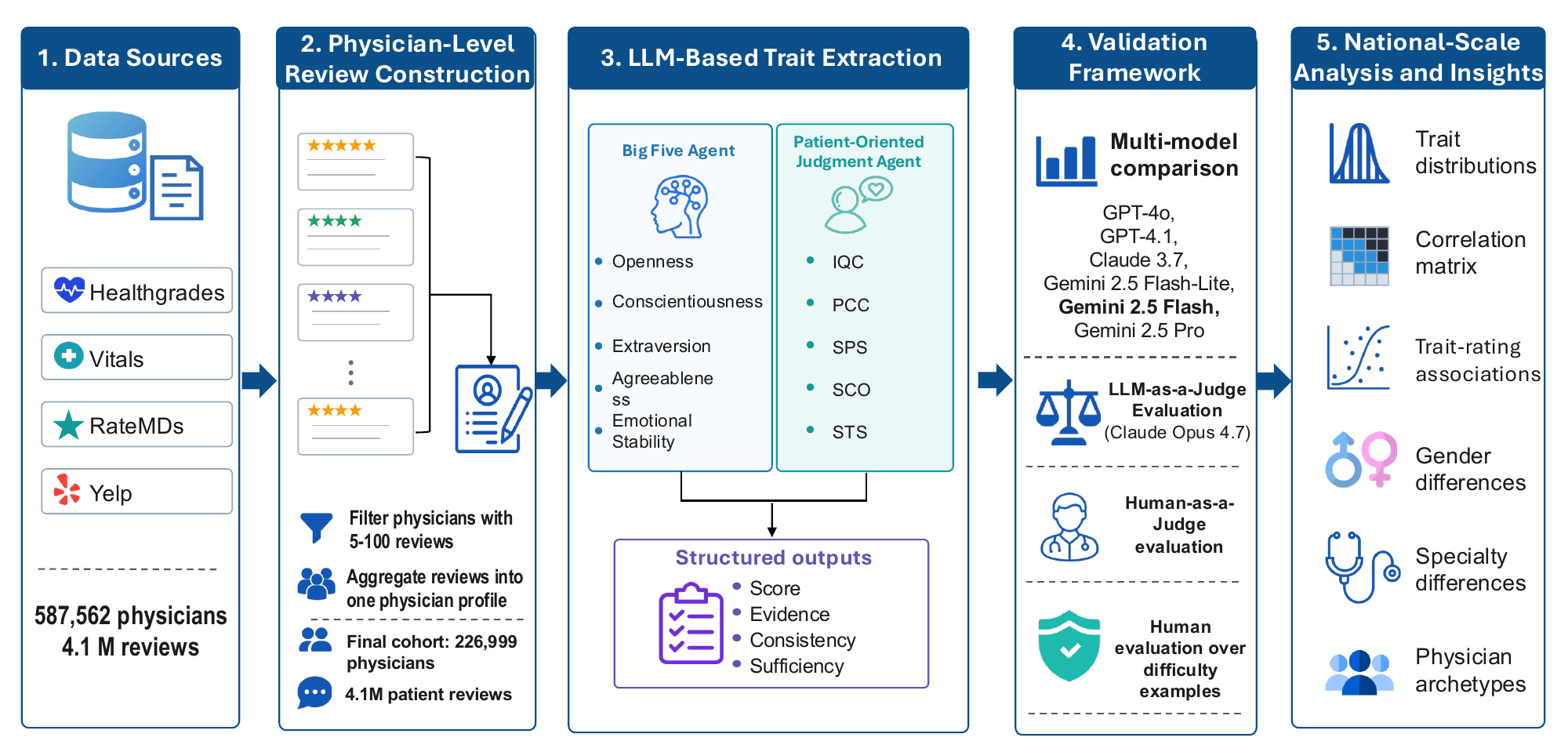}
\caption{
    Overview of the LLM-based pipeline for mapping patient-perceived physician traits at national scale. 
    \textbf{Stage 1 (Data Sources):} 8.69M reviews of 587{,}562 physicians from Healthgrades, Vitals, RateMDs, and Yelp. 
    \textbf{Stage 2 (Review Construction):} physicians with 5--100 reviews retained and aggregated, yielding 226{,}999 physicians with 4.1M reviews (\emph{Dataset Characteristics}).
    \textbf{Stage 3 (Trait Extraction):} two LLM agents extract five Big-Five-style and five patient-oriented traits with structured score, evidence, consistency, and sufficiency outputs (\emph{Automated LLM-based Trait Extraction}).
    \textbf{Stage 4 (Validation):} multi-model comparison and Human-as-a-Judge evaluation on 300-physician random and 200-case hard-case panels, across six LLMs with \textbf{Claude Opus 4.7}~\cite{claudeopus47_2026} as the cross-family judge (Table~\ref{tab:model_performance}).
    \textbf{Stage 5 (National-Scale Analysis):} trait distributions (\emph{Trait Distributions and Inter-Trait Structure}), gender and specialty (\emph{Subgroup Analysis}), archetypes (\emph{Cluster Analysis}), and trait--satisfaction (\emph{Trait-Rating Correlation}).
    Figure created by the authors using Microsoft PowerPoint.}
\label{fig:pipeline_overview}
\end{figure}

%
To map patient-perceived physician traits at national scale, we developed an LLM-based pipeline that extracts ten trait scores from millions of online patient reviews and evaluates them against multi-model and human-annotator benchmarks.
%
Figure~\ref{fig:pipeline_overview} summarizes the five-stage workflow (data sources, review aggregation, LLM trait extraction, validation, and national-scale analyses), each detailed in the subsection indicated.

\subsection{Dataset Characteristics and Scope}

%
%

We assembled a national corpus of 8.69 million online patient reviews of 587{,}562 U.S. physicians collected through August 2021 from four major rating platforms (Healthgrades, Vitals, RateMDs, and Yelp; Figure~\ref{fig:pipeline_overview}, Stage 1)~\cite{wang2022recency}.
%
We retained physicians with 5--100 reviews and aggregated each physician's reviews into a single profile to ensure adequate textual signal while maintaining representativeness.
%
The final analytic cohort comprised 226{,}999 physicians and 4.1 million reviews across all major medical specialties and geographic regions (Figure~\ref{fig:pipeline_overview}, Stage 2).
%
Dataset demographics, including gender, race, specialty, geographic region, mean review rating score, and review-count distributions, are reported in Supplementary Table~S1.

%
%

\subsection{Trait Targets and Definitions}

%
We extract ten subjective characteristics from online patient reviews.
%
These fall into two categories: five Big-Five-style interpersonal traits adapted from personality psychology, and five healthcare-specific patient-oriented subjective judgment traits.
%
We do not measure physician personality directly; we extract patient-perceived analogues from review narratives and call them patient-perceived traits throughout. 
%

%
The first category consists of five Big-Five-style interpersonal traits: openness, conscientiousness, extraversion, agreeableness, and emotional stability (the reverse-keyed Big Five ``neuroticism'').
%
The original Big Five traits are stable personality constructs in self-report data~\cite{widiger2019five}, and have been applied in healthcare to characterize physician interpersonal style and patient experience~\cite{morishita2025association}.
%
Our patient-perceived versions of these five traits are defined as follows.
%
Openness reflects a physician's intellectual curiosity, receptiveness to new approaches, and creative problem-solving.
%
Conscientiousness captures organization, reliability, and adherence to plans; patients often associate these with attentiveness and safety in care.
%
Extraversion encompasses sociability, assertiveness, and enthusiasm, which can shape a patient's experience of warmth and engagement during clinical encounters.
%
Agreeableness denotes cooperativeness, empathy, and emotional sensitivity; patients cite these in perceptions of kindness, compassion, and respectful communication.
%
Emotional stability (the reverse-keyed Big Five neuroticism) captures composure under pressure, which patients associate with confidence and clarity.
%
Lower scores reflect perceptions of stress reactivity or emotional instability.


\begin{table}[htbp]
\centering
\renewcommand{\arraystretch}{1.25} 

\begin{tabular}{
  >{\centering\arraybackslash}m{2cm}  
  >{\centering\arraybackslash}m{2.7cm}  
  >{\raggedright\arraybackslash}m{9cm} 
}
\toprule
\textbf{Category} & \textbf{Trait} & \textbf{Definition and Clinical Interpretation} \\
\midrule

\multirow{5}{*}{\shortstack{Big Five\\Personality}}
& Openness & Intellectual curiosity, receptiveness to new approaches, and creative problem-solving. \\
& Conscien-tiousness & Organization, reliability, and adherence to plans; patients often associate these with attentiveness and safety in care. \\
& Extraversion & Sociability, assertiveness, and enthusiasm; can shape a patient's experience of warmth and engagement during clinical encounters. \\
& Agreeableness & Cooperativeness, empathy, and emotional sensitivity; patients cite these in perceptions of kindness, compassion, and respectful communication. \\
& Emotional Stability (reverse-keyed Neuroticism) & Composure under pressure; patients associate higher scores with confidence and clarity, lower scores with stress reactivity or emotional instability. \\
\midrule

\multirow{5}{*}{\shortstack{Subjective\\Judgments}}
& IQC (Interpersonal Qualities \& Communication) & The physician's tone, empathy, clarity, listening, and respectfulness as inferred from the patient narrative. \\
& PCC (Perceived Clinical Competence) & The patient's impression of medical skill, judgment, and professionalism; whether the doctor is seen as knowledgeable, accurate, and effective. \\
& SPS (Sensitivity to Patient Satisfaction) & Whether the physician appears to care about the patient's preferences, comfort, and emotional needs. \\
& SCO (Sensitivity to Clinical Outcome) & Whether the physician is perceived to care about the patient's recovery, treatment success, or long-term health. \\
& STS (Social Trust Signals) & Indicators of trustworthiness such as being recommended by other patients or used repeatedly by families. \\
\bottomrule
\end{tabular}
\caption{Categories, Traits, and Clinical Interpretations of Physician Characteristics}
\label{table1}
\end{table}

%
The second category comprises five subjective judgment traits developed for interpreting online physician reviews~\cite{madanay2025physician, madanay2024classification}.
%
These traits reflect patient-facing perceptions beyond personality: behavioral performance, communication quality, and trustworthiness~\cite{hojat2011physicians}.
%
Interpersonal Qualities and Communication (IQC) refers to the physician's tone, empathy, clarity, listening, and respectfulness as inferred from the patient narrative.
%
Perceived Clinical Competence (PCC) assesses the patient's impression of medical skill, judgment, and professionalism: whether the doctor is seen as knowledgeable, accurate, and effective.
%
Sensitivity to Patient Satisfaction (SPS) reflects whether the physician appears to care about the patient's preferences, comfort, and emotional needs.
%
Sensitivity to Clinical Outcome (SCO) relates to whether the physician is perceived to care about the patient's recovery, treatment success, or long-term health.
%
Social Trust Signals (STS) capture indicators of trustworthiness such as being recommended by other patients or used repeatedly by families.
%

%
We selected these ten traits for their interpretability, frequency in patient narratives, and relevance to healthcare quality and patient-centered care.
%
Table~\ref{table1} summarizes the trait definitions.

%
%

\subsection{Automated LLM-based Trait Extraction Approach and Evaluation}

%
Stage 3 of the pipeline (Figure~\ref{fig:pipeline_overview}) uses two LLM agents that read each physician's aggregated reviews: one scores the five Big-Five-style interpersonal traits, the other scores the five patient-oriented judgment traits.
%
Both agents use chain-of-thought prompting~\cite{wei2022chain}, with per-trait Likert scores mapped to a 0--1 numeric value for downstream analysis (Figure~\ref{fig:pipeline_overview}, Stage 3).
%
For each trait, the LLM either records No Evidence when the reviews carry no relevant signal, or rates the trait on a five-level anchored Likert scale: Low, Low to Moderate, Moderate, Moderate to High, or High.
%
The same response includes a 2--3 sentence rationale grounded in patient text and two Low / Moderate / High ratings, one for consistency of the trait signal across reviews and one for evidence sufficiency.
%
The Big Five prompt assigns an ``expert psychologist'' role and lists the five patient-perceived Big Five traits; the full template is in \hyperref[sec:bigfive_prompt]{Supplementary Note~1}.
%
The SubFive prompt assigns an ``expert analyst'' role on patient perception, lists the five patient-oriented traits with one high-example and one low-example excerpt each, and includes IQC vs.\ SPS and PCC vs.\ SCO disambiguation notes; the full template is in \hyperref[sec:subfive_prompt]{Supplementary Note~2}.

%
We benchmark several candidate LLMs on this output schema, select one that is both accurate and cost-efficient at scale, and apply it to all 226{,}999 physicians.
%
Three layers exist in this setting: LLM-derived traits (what the LLM extracts from reviews), patient-perceived traits (what reviews encode about the physician), and physicians' real personality (an unobservable ground truth).
%
We assess the alignment between the LLM-derived and patient-perceived layers, not access to physicians' real personality, using a four-component framework (Figure~\ref{fig:pipeline_overview}, Stage 4): multi-model comparison, LLM-as-a-Judge with a cross-family judge, Human-as-a-Judge on 300 random physicians, and Human-as-a-Judge on 200 hard cases at the disagreement frontier.
%
Six LLMs spanning three families were scored on a 1{,}197-physician $\times$ 10-trait pool (11{,}970 cases): Gemini-2.5 Flash~\cite{gemini25flash2025}, Gemini-2.5 Pro~\cite{gemini25pro2025}, Gemini-2.5 Flash-Lite~\cite{gemini25flashlite2025}, GPT-4o~\cite{hurst2024gpt}, GPT-4.1~\cite{openai2024gpt41}, and Claude 3.7 Sonnet~\cite{claude37_2025}.
%
Claude Opus 4.7 served as the cross-family judge so that no candidate model judged itself.

\begin{table}[htbp]
\centering
\renewcommand{\arraystretch}{1.25}
\resizebox{\textwidth}{!}{%
\begin{tabular}{l c c c c c c c c c c c c}
\toprule
 & \multicolumn{4}{c}{\textbf{LLM-as-a-Judge}}
 & \multicolumn{4}{c}{\textbf{Human-as-a-Judge}}
 & \multicolumn{4}{c}{\textbf{Human-as-a-Judge (hard-case)}} \\
 & \multicolumn{4}{c}{\textit{$n = 1{,}197$}}
 & \multicolumn{4}{c}{\textit{$n = 300$}}
 & \multicolumn{4}{c}{\textit{$n = 200$ hard cases}} \\
\cmidrule(lr){2-5} \cmidrule(lr){6-9} \cmidrule(lr){10-13}
\textbf{Model}
  & \textbf{MAE} & \textbf{RMSE} & \textbf{High~\%} & \textbf{Low~\%}
  & \textbf{MAE} & \textbf{RMSE} & \textbf{High~\%} & \textbf{Low~\%}
  & \textbf{MAE} & \textbf{RMSE} & \textbf{High~\%} & \textbf{Low~\%} \\
\midrule
gemini-2.5-pro & 0.0845 & 0.1532 & 94.56\% & 75.57\% & 0.1851 & 0.2598 & 73.3\% & 87.2\% & 0.207 & 0.276 & 33.3\% & 10.0\% \\
gemini-2.5-flash & 0.0949 & 0.1654 & 90.94\% & 81.56\% & 0.1801 & 0.2625 & 78.5\% & 77.6\% & 0.244 & 0.313 & 14.3\% & 8.3\% \\
gpt-4.1 & 0.1102 & 0.1861 & 90.73\% & 89.72\% & 0.1493 & 0.2174 & 81.8\% & 71.1\% & 0.386 & 0.462 & 0.0\% & 43.8\% \\
claude-3.7-sonnet & 0.1037 & 0.1780 & 87.41\% & 86.65\% & 0.1577 & 0.2389 & 80.2\% & 66.7\% & 0.230 & 0.305 & 42.9\% & 5.6\% \\
gpt-4o & 0.1256 & 0.2066 & 96.24\% & 87.14\% & 0.1552 & 0.2241 & 80.7\% & 69.1\% & 0.279 & 0.362 & 71.4\% & 11.1\% \\
gemini-2.5-flash-lite & 0.1307 & 0.2163 & 92.18\% & 83.76\% & 0.1837 & 0.2685 & 88.1\% & 65.2\% & 0.310 & 0.376 & 0.0\% & 31.2\% \\
\bottomrule
\end{tabular}%
}
\caption{Model performance under three evaluation regimes against the Claude Opus 4.7 cross-family judge (LLM-as-a-Judge, 1{,}197 physicians) and human raters (random 300; hard-case 200 drawn from the disagreement frontier, Supplementary Section~S5.4). MAE/RMSE are on the 0--1 ordinal score (Low=0, Low-to-Moderate=0.25, Moderate=0.5, Moderate-to-High=0.75, High=1); High~\% / Low~\% are exact-bucket match rates on items the judge scored High / Low.}
\label{tab:model_performance}
\end{table}

%
Model performance varied across the three evaluation panels (Table~\ref{tab:model_performance}); on the LLM-judge panel ($n = 1{,}197$), the top tier of Gemini-2.5 Pro (MAE = 0.0845), Gemini-2.5 Flash (0.0949), Claude 3.7 Sonnet (0.1037), and GPT-4.1 (0.1102) clusters within $\sim$0.03 MAE.
%
GPT-4o (0.1256) and Gemini-2.5 Flash-Lite (0.1307) trail by 0.015--0.020 MAE.
%
RMSE follows the same ordering (top tier 0.153--0.186 vs.\ bottom tier 0.207--0.216), and the six candidates land within 87--96\% on items the Opus judge scored High and 76--90\% on items it scored Low (Table~\ref{tab:model_performance}, left panel).
%
The Opus judge separates a tight 4-model top tier from a clearly weaker 2-model tail, and the within-family Gemini-2.5 Pro judge reproduces the same ordering (overall Spearman $\rho = 0.943$; \hyperref[sec:cross_family_judge]{Supplementary Note~5.5}), so the cross-family separation is stable across both judge families and model families.
%

%
On the random panel ($n = 300$; Table~\ref{tab:model_performance}, middle), the six models cluster in a tight MAE band of 0.149--0.185 against human raters.
%
RMSE follows the same compression (0.217--0.269), with exact-bucket match rates of 73--88\% on items humans scored High and 65--87\% on items they scored Low.
%
At population scale the LLM scores match or exceed human inter-rater consistency, especially for the patient-oriented judgments.
%

%
The hard-case panel ($n = 200$) was drawn from the lowest $\sim$1.7\% of the 11{,}970-case judge dataset by cross-model agreement (20 per trait; each annotated by one of 5 RAs; Figure~\ref{fig:pipeline_overview}, Stage 4).
%
On these cases the ordering shifts: Gemini-2.5 Pro leads (MAE = 0.207) and GPT-4.1 trails (MAE = 0.386); the six candidates span 0.207--0.386 MAE (\hyperref[sec:hard_case_validation]{Supplementary Note~5.4}).
%
At the disagreement frontier no single candidate is reliable on its own, but the cross-family Opus synthesis stays closer to human judgment than any single LLM, validating it as our evaluation benchmark while remaining too expensive for full-cohort extraction.
%

%
Across the three panels, Gemini-2.5 Flash~\cite{gemini25flash2025} sits in the top tier on accuracy (MAE = 0.0949 under the Opus judge on 11{,}970 cases) at low per-token cost.
%
We therefore apply Gemini-2.5 Flash to all 226{,}999 physicians for full-cohort trait extraction.

\subsection{Trait Distributions and Inter-Trait Structure}

%
%

%
We applied Gemini-2.5 Flash to all 226{,}999 physicians to extract patient-perceived trait scores.
%
We first describe how the ten trait scores are distributed across physicians.
%
We then test whether the ten traits remain distinguishable from each other once review rating scores are held fixed.
%
Extraction-quality checks (consistency, sufficiency, and the raw vs.\ sentiment-controlled U-shape) are in \hyperref[sec:quality_uShape]{Supplementary Note~5.3}.

\begin{figure}[htbp]
\centering
\includegraphics[width=\textwidth]{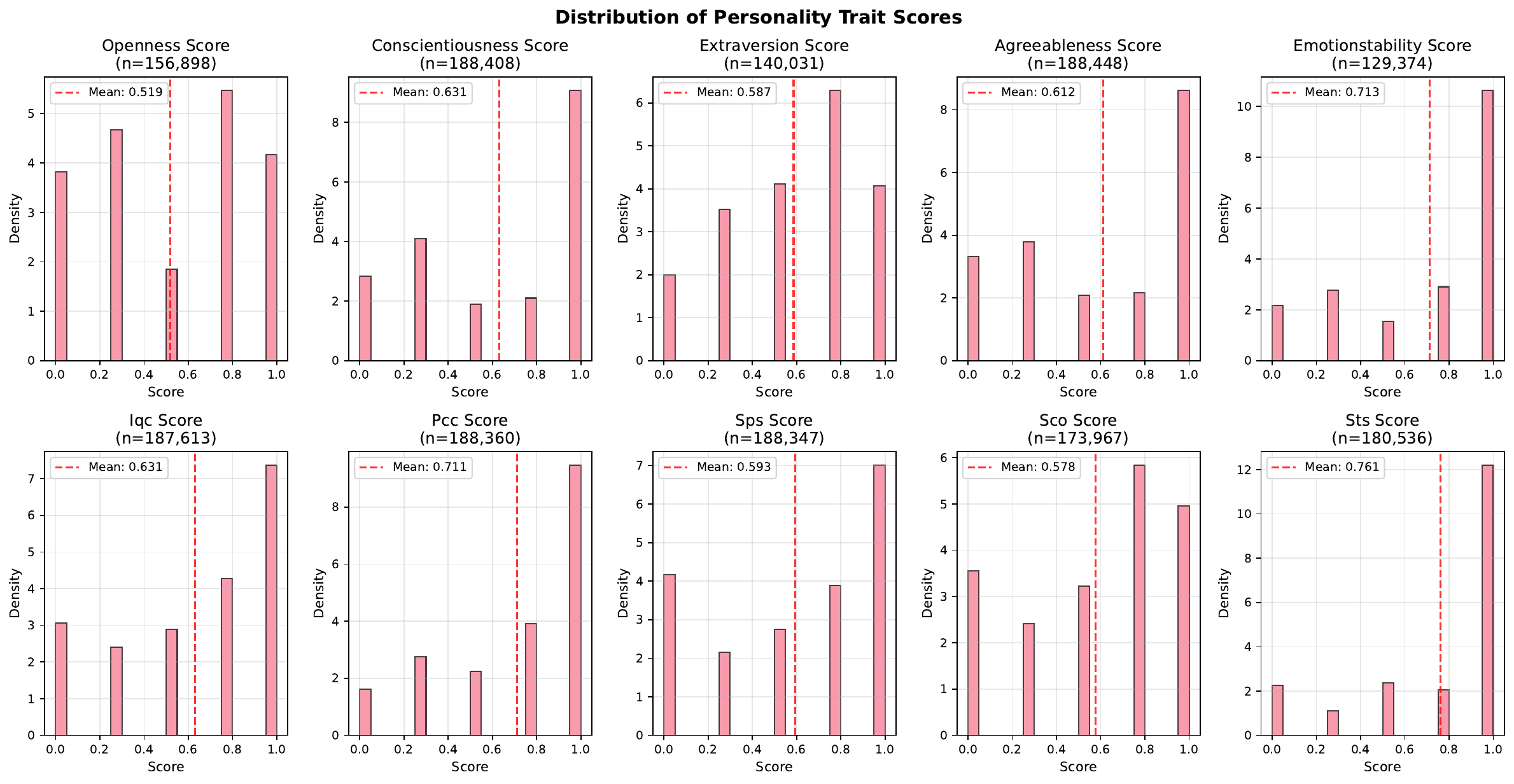}
\caption{Distribution of patient-perceived trait scores. Each subplot shows the kernel density estimate for one of the 10 traits on a 0--1 scale. Figure created by the authors using Matplotlib (Python).}
\label{fig:trait_distributions}
\end{figure}

%
Trait-score means span 0.519 (Openness) to 0.761 (Social Trust Signals), with subjective-judgment traits running higher than Big-Five-style traits (Figure~\ref{fig:trait_distributions}).
%
Among the Big Five traits, Emotional Stability had the highest mean ($M = 0.713$, 95\% CI [0.711, 0.715]), followed by Conscientiousness (0.631 [0.629, 0.633]) and Agreeableness (0.612 [0.610, 0.614]).
%
Extraversion was intermediate (0.587 [0.585, 0.588]) and Openness lowest (0.519 [0.517, 0.520]).
%
Among subjective judgments, Social Trust Signals led (STS; $M = 0.761$, 95\% CI [0.759, 0.762]), followed by Perceived Clinical Competence (PCC; 0.711 [0.709, 0.713]) and Interpersonal Qualities \& Communication (IQC; 0.631 [0.630, 0.633]).
%
Sensitivity to Patient Satisfaction (SPS; $M = 0.593$, 95\% CI [0.591, 0.594]) and Sensitivity to Clinical Outcome (SCO; $M = 0.578$, 95\% CI [0.576, 0.580]) had the lowest means in this group.
%
All trait-mean CIs are 1{,}000-resample bootstrap 95\% intervals.

\begin{figure}[htbp]
\centering
\includegraphics[width=\textwidth]{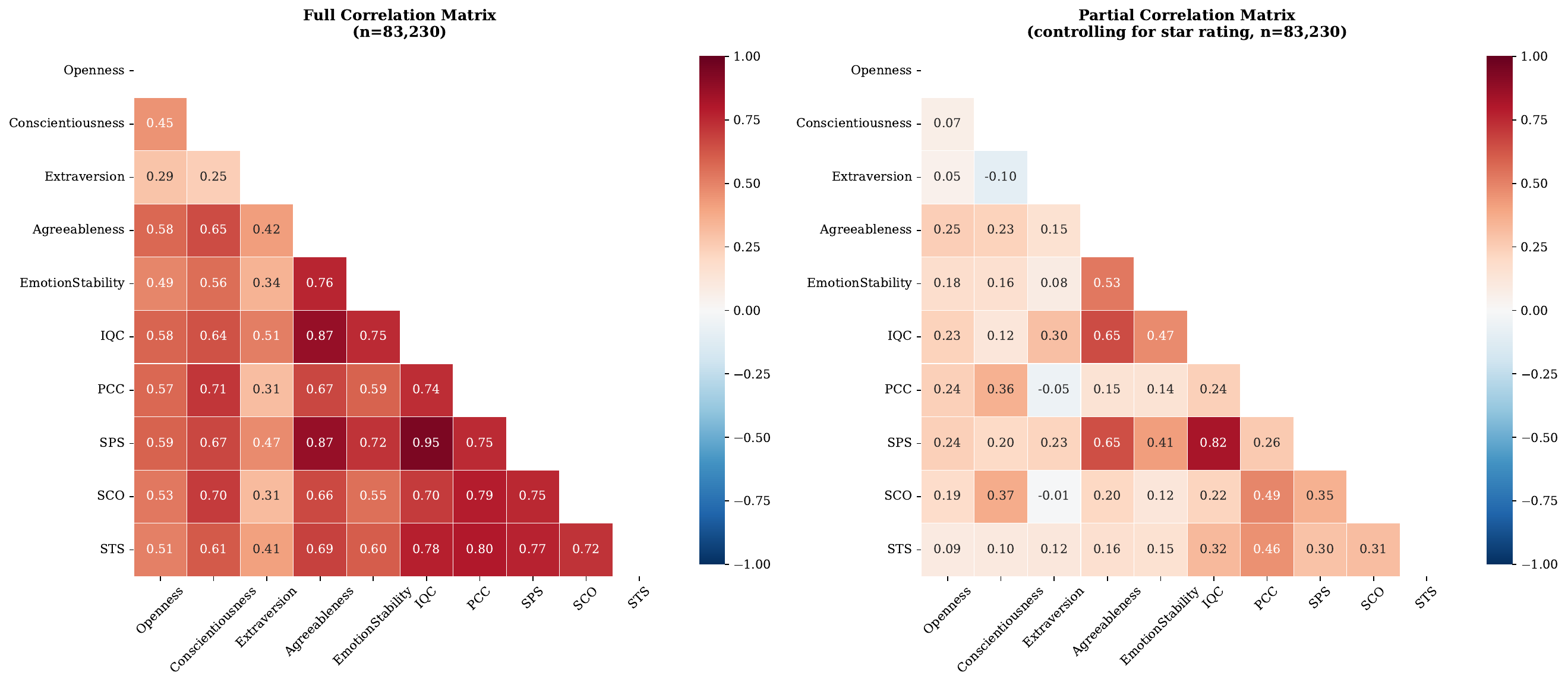}
\caption{Trait--trait correlations: raw versus sentiment-controlled heatmaps. \textbf{Left:} zero-order Pearson correlations among the 10 patient-perceived traits (mean inter-trait $r = 0.61$; IQC--SPS $r = 0.928$). \textbf{Right:} partial correlations after controlling for per-physician mean review rating score as a sentiment proxy; the mean inter-trait $r$ drops to $0.25$, yet all 45 of 45 trait pairs remain statistically significant ($p < 0.001$, Bonferroni-corrected), indicating discriminant structure beyond shared sentiment variance. Figure created by the authors using Matplotlib (Python).}
\label{fig:trait_correlations}
\end{figure}

%
Once review rating scores are partialled out, all 45 of 45 trait pairs remain correlated (mean partial $r = 0.25$, $p < 0.001$ Bonferroni-corrected; Figure~\ref{fig:trait_correlations} right panel).
%
This residual inter-trait $r$ is in the same moderate range as Big Five intercorrelations reported in self-report studies~\cite{van2010general,deyoung2006higher,musek2007general}.
%
Our higher raw mean ($r = 0.61$) reflects patient-perceived traits extracted from a single review narrative rather than self-report across multiple instruments.
%
Even the most overlapping pair, IQC and SPS, separates systematically by specialty (gap +0.056 in Family Medicine to +0.020 in Surgery; ANOVA $F = 51.45$, $p < 0.001$).
%
From the above analysis, the ten traits, although strongly intercorrelated, remain distinguishable from each other once review rating scores are held constant.
%

%
%

\subsection{Trait-Rating Correlation}

\begin{figure}[htbp]
\centering
\includegraphics[width=\textwidth]{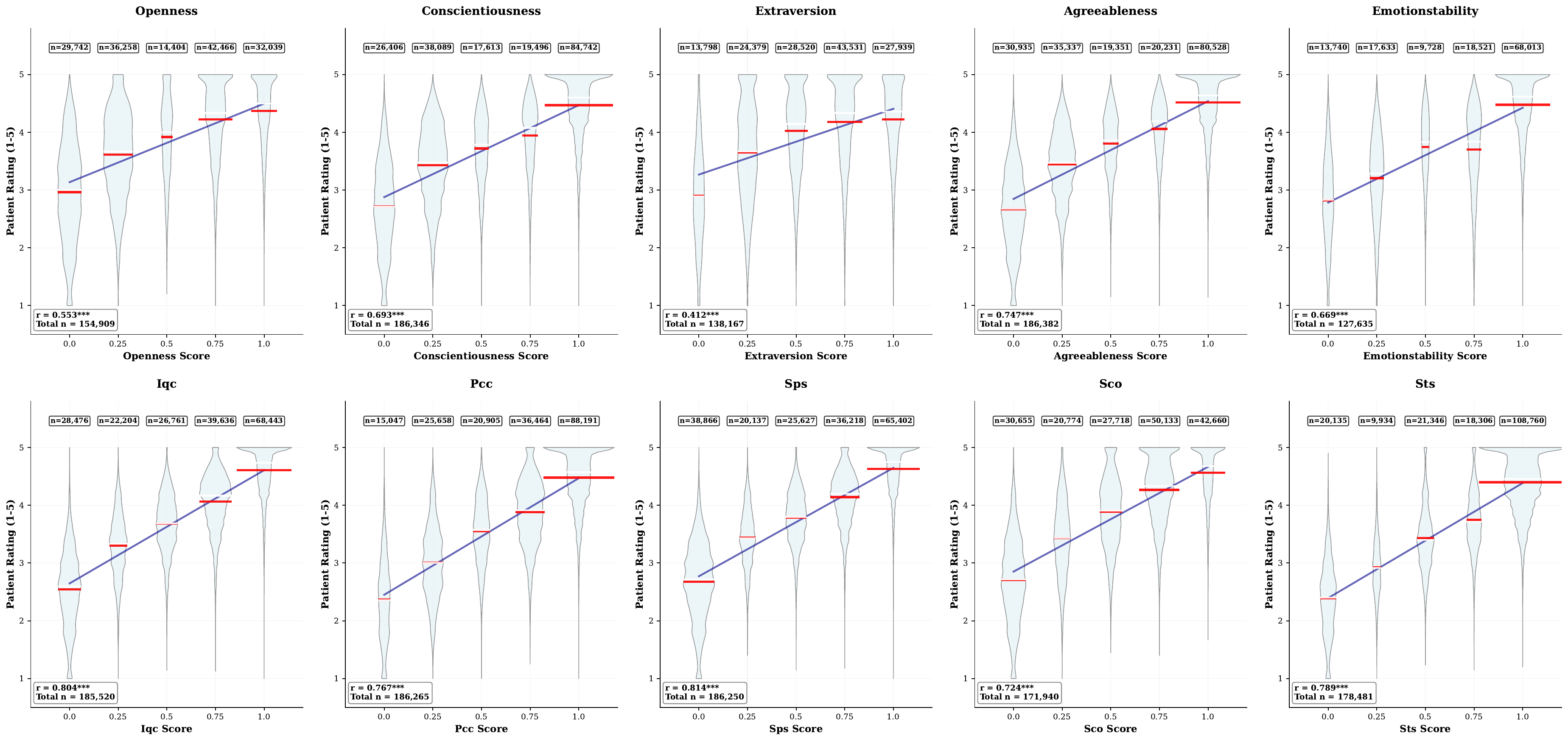}
\caption{Trait scores and the review rating score. Area-scaled violin plots show review rating score distributions for discrete trait levels (0, 0.25, 0.5, 0.75, 1.0); violin width is proportional to sample size. The review rating score rises monotonically with trait score across all 10 traits, most steeply for patient-oriented subjective judgments (SPS, IQC, STS). Figure created by the authors using Matplotlib (Python).}
\label{fig:trait_satisfaction}
\end{figure}

%
In the previous subsection, we established that the ten traits remain distinguishable from each other.
%
We now ask whether each trait retains variance independent of review rating scores.
%
Trait--satisfaction correlations reach $r = 0.81$ (Figure~\ref{fig:trait_satisfaction}).
%
Such correlations are expected under a patient-perceived framing since both signals come from the same review narrative; the operative test is whether they survive controlling for review rating scores.
%

%
The area-scaled violin plots show monotonic increases in the review rating score as trait scores rise from 0 to 1, with the largest effects for patient-oriented subjective judgments.
%
Sensitivity to Patient Satisfaction (SPS) had the strongest correlation ($r = 0.814$, 95\% CI [0.812, 0.815], $n = 186{,}250$), followed by Interpersonal Qualities \& Communication ($r = 0.804$ [0.802, 0.806]) and Social Trust Signals ($r = 0.789$ [0.787, 0.790]); all $p<0.001$.
%
Partial correlations controlling for review rating scores remain non-zero across all ten traits, with Social Trust Signals ($r_{\text{partial}} = 0.28$) and Conscientiousness ($0.21$) retaining the most independent signal (attenuation range 66--96\%).
%
All trait--satisfaction CIs are 1{,}000-resample bootstrap 95\% intervals.

%
Among the Big Five, Agreeableness ($r = 0.747$, 95\% CI [0.745, 0.749]) and Conscientiousness ($r = 0.693$ [0.691, 0.696]) had the strongest associations with the review rating score, and Extraversion the weakest ($r = 0.412$ [0.407, 0.416]).
%
Violin shapes tighten at higher trait levels: physicians scoring 1.0 cluster around 4--5 stars while those scoring 0 spread broadly toward lower ratings.
%
A multiple regression of standardized trait scores on the review rating score (10 predictors, $n = 83{,}230$, $R^2 = 0.803$) showed that these traits independently contribute to the review rating score.
%
Social Trust Signals was the strongest standardized predictor ($\beta = 0.213$, 95\% CI [0.207, 0.219]), followed by Interpersonal Qualities \& Communication ($\beta = 0.134$ [0.124, 0.144]) and Conscientiousness ($\beta = 0.132$ [0.128, 0.136]).
%
All 10 predictors' CIs are bounded away from zero, down to Emotional Stability ($\beta = 0.015$, 95\% CI [0.010, 0.020]).
%
From the above analysis, trait scores correlate strongly with review rating scores yet each retains substantial independent variance.
%
%
%

\subsection{Subgroup Analysis, Gender \& Specialty}

\begin{figure}[htbp]
\centering
\includegraphics[width=\textwidth]{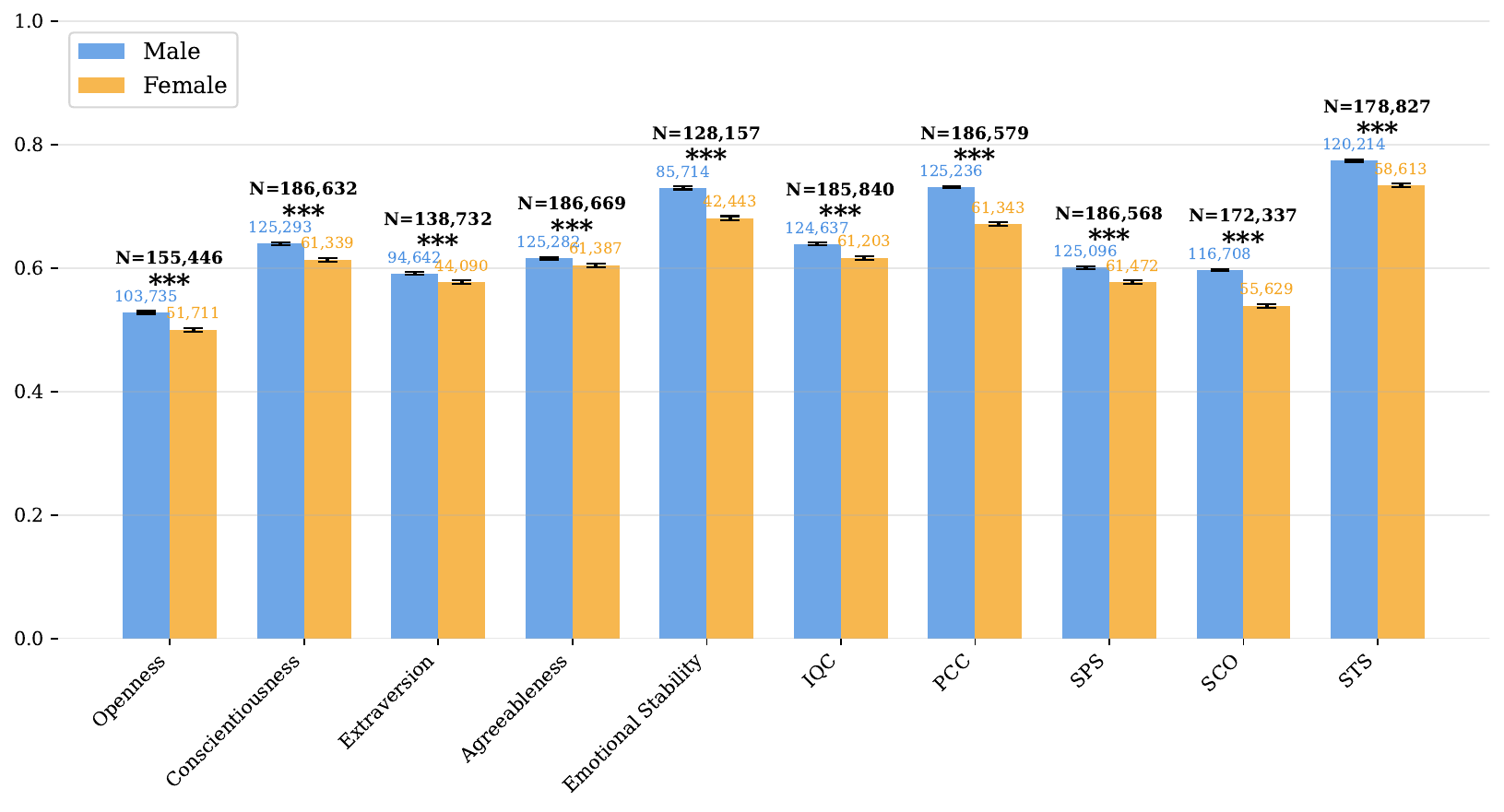}
\caption{Gender differences in physician trait scores. Male physicians receive higher ratings across all 10 traits, with largest differences in patient-oriented subjective judgments (PCC, SCO, STS) rather than personality traits. Error bars represent 95\% confidence intervals calculated from standard errors. Statistical significance was determined using independent-samples t-tests. Sample sizes vary by trait (n$_{\text{male}}$ = 85,714--125,293; n$_{\text{female}}$ = 42,443--61,472; total n = 128,157--186,669). Figure created by the authors using Matplotlib (Python).}
\label{fig:gender_differences}
\end{figure} 

%
Across 128,157 to 186,669 physicians per trait, male physicians received higher ratings than female physicians on all ten traits ($p<0.001$; Figure~\ref{fig:gender_differences}).
%
All ten gaps pointed in the same direction across personality traits and subjective judgments (Cohen's $d$ = 0.029 to 0.177).
%
Gaps were largest in PCC, SCO, Emotional Stability, and STS; the other six traits all had $|d| < 0.08$.
%
Perceived Clinical Competence had the largest disparity ($|d| = 0.177$ [0.167, 0.186]), followed by Sensitivity to Clinical Outcome (0.164 [0.154, 0.175]), Emotional Stability (0.136 [0.124, 0.147]), Social Trust Signals (0.116 [0.107, 0.127]), and Openness (0.078 [0.068, 0.088]).
%
Agreeableness ($|d| = 0.029$, 95\% CI [0.019, 0.038]) and Extraversion ($|d| = 0.046$, 95\% CI [0.034, 0.057]) showed the smallest gender differences.
%
All Cohen's $d$ values use 1{,}000-resample bootstrap 95\% intervals.

%
All ten traits move in the same direction, with female physicians scoring lower on every trait.
%
The consistency of this direction across traits is more notable than the magnitude of any single trait.
%
At our sample size the differences reach statistical significance even at small effect sizes (Cohen's $d$ = 0.029--0.177), so practical significance at the individual level may be limited.
%
The pattern is consistent with prior work on physician ratings in patient evaluations~\cite{madanay2025physician}.
%
Future work should diagnose whether this gender pattern reflects review bias, workforce composition, or both, and how to prevent these biases from shaping workforce decisions.

\begin{figure}[htbp]
\centering
\begin{subfigure}[b]{0.48\textwidth}
\centering
\includegraphics[width=\textwidth]{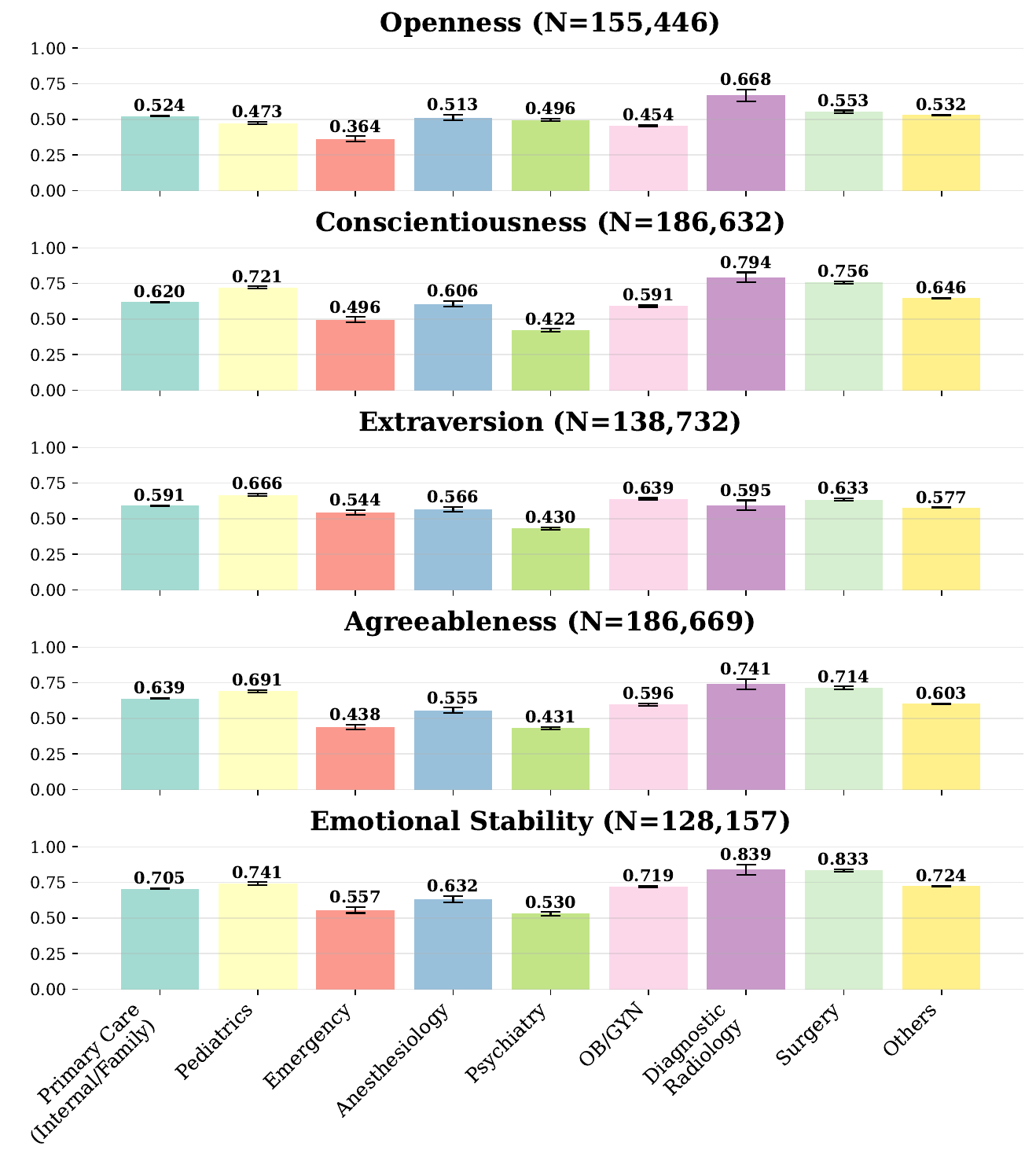}
\caption{BigFive personality traits by specialty}
\label{fig:specialty_bigfive}
\end{subfigure}
\hfill
\begin{subfigure}[b]{0.48\textwidth}
\centering
\includegraphics[width=\textwidth]{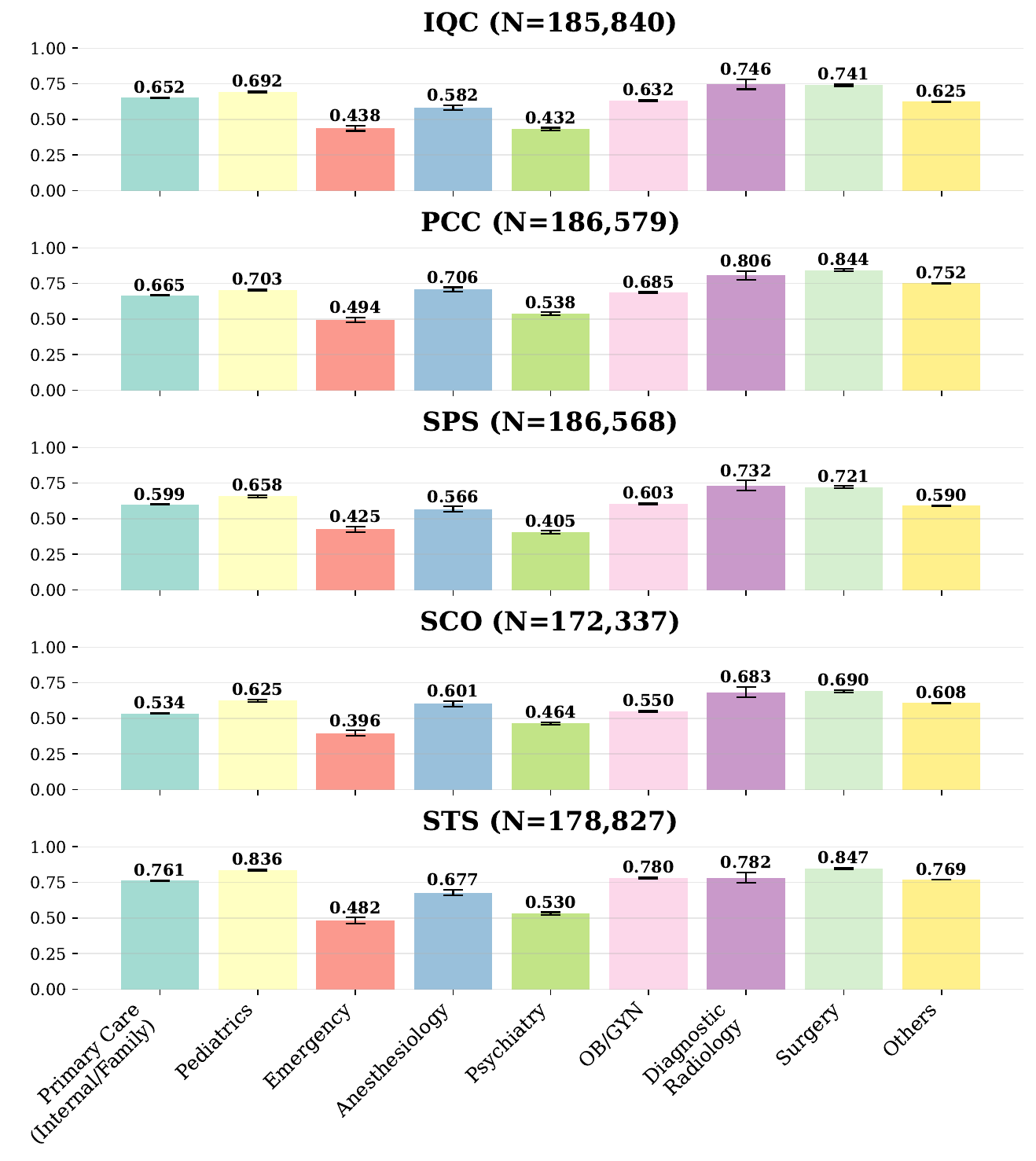}
\caption{Professional competencies by specialty}
\label{fig:specialty_competencies}
\end{subfigure}
\caption{Specialty differences in physician traits. (a) Big-Five personality trait scores by medical specialty show distinct patterns; surgical specialties score higher on Conscientiousness and Emotional Stability. (b) Professional competency measures show specialty-specific strengths; surgical specialties score highest on Perceived Clinical Competence, while patient-oriented specialties score higher on Interpersonal Qualities \& Communication. Figure created by the authors using Matplotlib (Python).}
\label{fig:specialty_analysis}
\end{figure} 

%
Trait distributions varied significantly across nine major medical specialties and an aggregated ``Others'' category (Figure~\ref{fig:specialty_analysis}; total $N$ = 100,181 to 146,787 per trait), reflecting both selection effects and the distinct demands of different fields~\cite{daskivich2018differences}.
%
Bootstrap omnibus $\eta^2$ (one-way ANOVA, 1{,}000 resamples) ranged from $\eta^2 = 0.012$ (Openness, 95\% CI [0.011, 0.014]) to $\eta^2 = 0.053$ (PCC, 95\% CI [0.050, 0.055]).
%
Patient-oriented traits (PCC, STS, SCO) carried the largest specialty-attributable variance; Openness the smallest.
%

%
Surgical specialties had the highest scores across nearly all domains.
%
Among personality traits and competencies, Diagnostic Radiology ranked highest in Conscientiousness (0.794) and Emotional Stability (0.839) (Figure~\ref{fig:specialty_bigfive}), and in PCC (0.896), SCO (0.683), and STS (0.782) (Figure~\ref{fig:specialty_competencies}).
%
Surgery, by contrast, led on the two patient-oriented measures, SPS (0.721) and IQC (0.741).
%
The Diagnostic Radiology pattern likely reflects review-selection bias: patients writing radiology reviews typically had direct consultative or interventional contact, yielding a positively selected sample.
%

%
Emergency Medicine and Psychiatry physicians scored lowest across multiple domains.
%
Emergency Medicine scored low on Openness (0.364), Conscientiousness (0.496), and Social Trust Signals (0.482).
%
These low scores are consistent with the acute, high-stress nature of Emergency Medicine encounters and the limited relationship-building they allow.
%
Psychiatry scored low on Conscientiousness (0.422), Extraversion (0.430), and Social Trust Signals (0.530).
%
These low scores likely reflect contextual factors (mental-health stigma, encounter complexity, condition difficulty) more than intrinsic physician traits.
%
Between these extremes, primary care specialties had moderate profiles, with Pediatrics scoring higher on Extraversion (0.666) and SPS (0.658) than Internal Medicine.

%
These specialty differences likely reflect several bidirectional mechanisms.
%
First, patient self-selection: individuals with particular conditions or personalities gravitate toward certain specialties, which can create more difficult evaluation contexts (notably Psychiatry).
%
Second, prolonged exposure to particular patient populations and clinical scenarios may shape physician traits and professional behaviors over time.
%
Third, structural factors such as time constraints in Emergency Medicine and stigma in Psychiatry influence interaction quality.
%
Patient interaction differs sharply across specialties, and these differences shape what patients are positioned to observe and rate.
%
LLM-derived trait scores used for evaluation or workforce policy should therefore be calibrated to practice context rather than applied uniformly across specialties.

\subsection{Cluster Analysis, Latent Profiles}

\begin{figure}[htbp]
\centering
\begin{subfigure}[b]{0.48\textwidth}
\centering
\includegraphics[width=\textwidth]{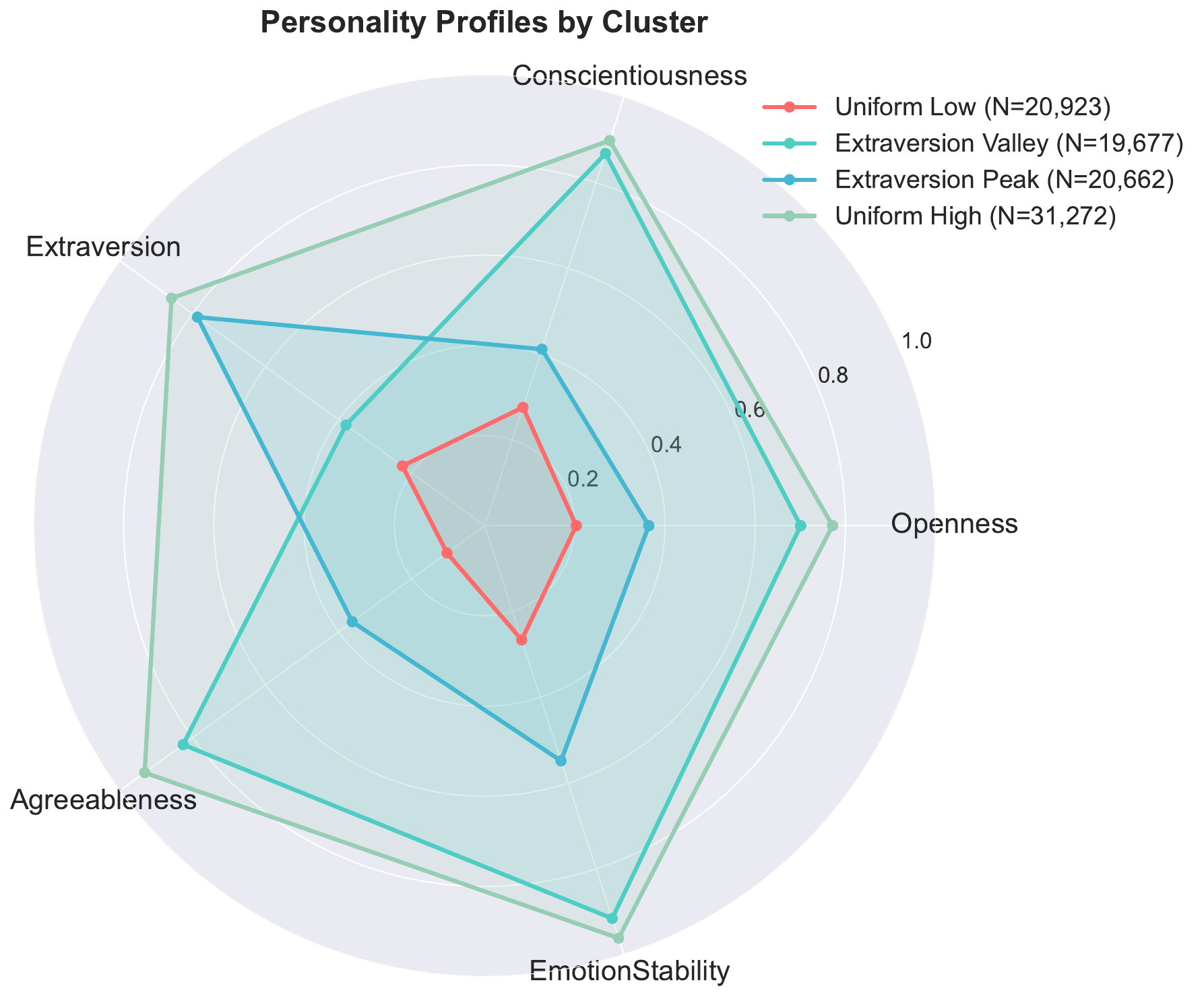}
\caption{BigFive personality profiles}
\label{fig:personality_clusters}
\end{subfigure}
\hfill
\begin{subfigure}[b]{0.48\textwidth}
\centering
\includegraphics[width=\textwidth]{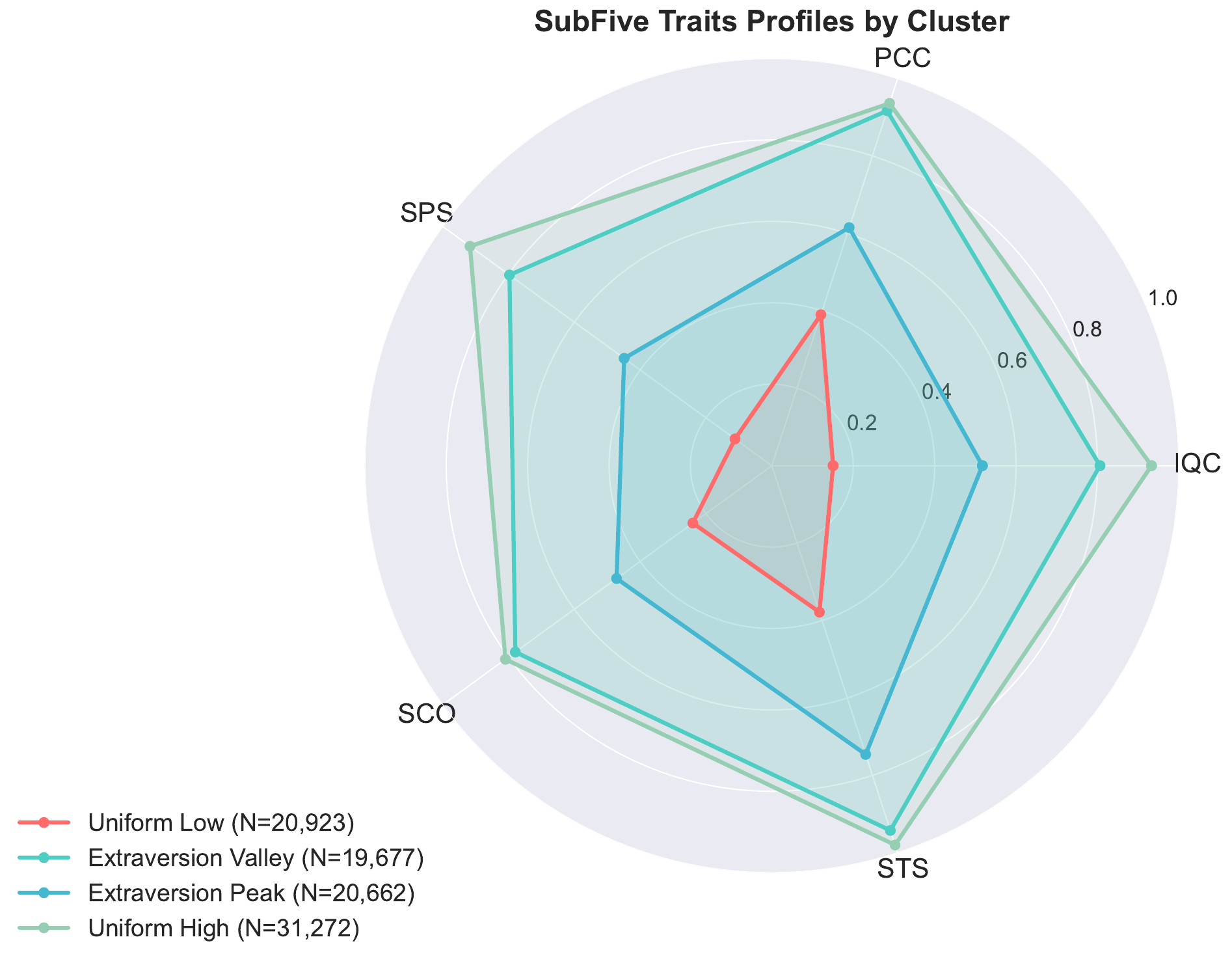}
\caption{Patient perception profiles}
\label{fig:patient_perception_clusters}
\end{subfigure}
\caption{Physician archetypes from K-means clustering. (a) Four Big Five profiles: Uniform High (high all), Uniform Low (low all), Extraversion Peak (high Extraversion only), and Extraversion Valley (high except Extraversion). Labels describe score patterns. (b) The same four archetypes show distinct profiles in patient-perceived traits, paralleling their Big Five patterns. Figure created by the authors using Matplotlib (Python).}
\label{fig:clustering_profiles}
\end{figure} 

%
Beyond the gender and specialty subgroups examined above, physicians may cluster naturally by their full trait profile.
%
K-means clustering ($k = 4$, chosen by elbow method; \hyperref[sec:clustering_validation]{Supplementary Note~4}) on the ten trait scores yielded four physician archetypes (Figure~\ref{fig:clustering_profiles}).

%
The ``Uniform High'' archetype (33.8\% of physicians) had high Big Five scores (Figure~\ref{fig:personality_clusters}): Emotional Stability 0.962, Agreeableness 0.931, Conscientiousness 0.898, Extraversion 0.858, Openness 0.772.
%
The ``Uniform Low'' archetype (22.6\% of physicians) had low Big Five scores: Agreeableness 0.103, Openness 0.203, Extraversion 0.225, Emotional Stability 0.266, Conscientiousness 0.276.
%
The ``Extraversion Peak'' archetype (22.3\% of physicians) had high Extraversion (0.786) and lower scores elsewhere: Agreeableness 0.363, Conscientiousness 0.411, Emotional Stability 0.549, Openness 0.364.
%
High Extraversion alone does not track with strong patient-rated clinical scores in this archetype.
%
The ``Extraversion Valley'' archetype (21.3\% of physicians) had strong Big Five scores (Emotional Stability 0.916, Conscientiousness 0.868, Agreeableness 0.826, Openness 0.701) despite low Extraversion (0.380).
%
The same archetype ordering held across Big Five and patient-perceived traits (Figure~\ref{fig:patient_perception_clusters}; Kruskal-Wallis $H$ = 46{,}153--74{,}009, all $p < 0.0001$).
%

%
To test whether archetypes differ in shape rather than level, we ran a one-way ANOVA on z-normalized (ipsatized) trait profiles within each cluster (Supplementary Figure~S5).
%
Clusters differed in shape, not just level (Conscientiousness $\eta^2 = 0.607$, Agreeableness 0.291, Extraversion 0.258, Emotional Stability 0.091, Openness 0.052; all $p < 0.001$).
%
Conscientiousness, not Extraversion, drives the largest shape difference, contrary to a level-only model.
%
For example, Extraversion Valley ($E = 0.380$) and Extraversion Peak ($E = 0.786$) show inverted shape profiles, not scaled-level differences.
%

%
Cluster assignments are highly stable: across 200 bootstrap resamples (80\% sample fraction, $k = 4$), the mean Adjusted Rand Index is 0.998 (95\% CI [0.988, 1.000]).
%
Boundary physicians (silhouette $< 0.1$) account for 15.4\% of the cohort.

%
Different personality configurations track with distinct patient interaction patterns.
%
These archetypes characterize physician profiles in the population.
%
Future work should validate these archetypes in independent clinical samples and develop the ethical framework needed for individual-level uses such as matching or credentialing.

\section{Discussion}

%
%

%
We applied large language models to extract patient-perceived physician trait scores from online reviews at national scale.
%
Our dataset comprised 4.1 million reviews for 226{,}999 U.S. physicians.
%
Throughout, our claims concern LLM-derived trait scores and the patient-perceived traits in reviews, not physicians' underlying personality.
%
Our LLM-extracted scores agree across models and with an independent human-annotation panel.
%
The extracted traits carry information about overall patient satisfaction beyond shared review sentiment, and show distinct patterns across demographics, specialties, and practice contexts.

%
Our multi-model evaluation supports LLM-based trait extraction in healthcare review text.
%
Cross-model consistency (Spearman $\rho$ = 0.943) and human-LLM agreement (MAE = 0.149--0.185 across six models on the random panel) together suggest trait constructs that are stable across raters~\cite{peters2024large}.
%
Partial-correlation analysis shows that the signal extends beyond shared sentiment at two levels: trait--satisfaction correlations attenuate by 66--96\% yet remain significant, and all 45 inter-trait pairs survive partialling star rating.
%
The dual framework combines Big Five personality traits with patient-oriented subjective judgments, linking psychological theory to clinical practice.
%
Within the subjective judgments, IQC and SPS overlap at the individual level ($r = 0.928$) but separate at the specialty level, so we treat them as distinct constructs.
%
The framework may inform healthcare quality measurement while preserving the patient perspective encoded in narrative feedback.
%

%
We identify four physician archetypes: Uniform High, Uniform Low, Extraversion Peak, and Extraversion Valley.
%
These archetypes may inform future research on physician interpersonal style and patient experience; individual-level use requires independent clinical validation and ethical review.
%
The Uniform High archetype covers 33.8\% of physicians and the Uniform Low archetype 22.6\%, indicating heterogeneity that single-rating systems do not capture.
%
The Extraversion Peak archetype (22.3\% of physicians) shows that high extraversion alone does not translate to higher patient satisfaction.
%
Interpersonal warmth alone is not sufficient for patient-perceived quality; empathy, reliability, and clinical competence are the main drivers.
%

%
The demographic disparities here, with male physicians receiving higher ratings across all ten traits, are consistent with documented biases in patient evaluation systems~\cite{fitzgerald2014physicians} and suggest patient-satisfaction metrics may disadvantage diverse physicians.
%
The specialty analysis suggests that physician trait patterns reflect both selection effects and adaptation to practice contexts.
%
Lower ratings for Emergency Medicine and Psychiatry physicians likely reflect the contexts these specialties operate in.
%
Emergency Medicine involves acute, high-stress encounters with limited relationship-building opportunities~\cite{Zhang2020burnout}.
%
Mental-health stigma, emotional complexity in psychiatric encounters, and the occupational stigma psychiatrists themselves face~\cite{Shi2023PsychiatristStigma} are contextual factors that outweigh intrinsic physician traits.
%
These patterns may inform specialty-specific quality metrics and context-adjusted assessments that account for the patient populations each specialty serves, rather than uniform evaluation standards.
%

%
We applied this pipeline to 226{,}999 physicians, showing that automated trait extraction works at national scale.
%
The pipeline is model-agnostic, and its evidence-anchored scores can be audited by clinical experts~\cite{rao2023can, peters2024large}.
%
Fair, generalizable use will require cautious interpretation and integration with complementary data sources.
%
Automated extraction can also propagate biases present in training data or patient-review patterns~\cite{champagne2021physician}.
%
Responsible use of LLM-derived trait scores in clinical practice will therefore require ongoing monitoring and validation by clinical experts before any individual-level application~\cite{minaee2024large, busch2025current}.
%

%
%
%

Our study has several limitations.
%
First, our pipeline measures patient-perceived traits, not psychometrically validated personality: the scores reflect how patients describe a clinical encounter, which is shaped by outcomes, expectations, and who chooses to write a review, rather than a physician's stable disposition.
%
We do not compare LLM scores against self-report instruments (NEO-PI-R, BFI), so human--LLM agreement is a consistency check rather than external ground-truth validation, and whether patient-perceived traits track underlying personality is left for future work using paired self-report data~\cite{youyou2015computer}.
%
Second, trait scores share variance with the review rating score; partial correlations controlling for the rating show that each trait retains signal beyond sentiment, but unconditional correlations should be read with this overlap in mind.
%
Third, our cohort contains only physicians with enough publicly posted reviews, who may differ from non-reviewed physicians in patient mix, geography, and online visibility, so our findings generalize to publicly reviewed physicians rather than the full physician population, and the cross-sectional design limits inference about trait stability over time~\cite{mcgrath2018validity,hanauer2014public,Roberts2006PersonalityChange}.
%
Fourth, human validation rested on a random panel and a disagreement-targeted hard-case panel that probes the cases where the models diverge most (\hyperref[sec:hard_case_validation]{Supplementary Note~5.4}); because annotation at this scale is costly and difficult to organize, we concentrated it on these regimes rather than pursuing exhaustive coverage, so it supports aggregate research conclusions rather than per-physician judgments~\cite{champagne2021physician,choo2017damned}.
%
The cross-family judge mitigates within-family bias in the LLM-as-a-Judge panel but cannot fully eliminate it, and broader psychometric validation remains future work.
%

%
%
%

%
Beyond these methodological limitations, profiling physician traits at national scale raises ethical and societal questions.
%
Our data and code were held in a HIPAA-compliant institutional environment for secondary analysis of publicly available data, and collection followed each platform's rate-limiting and access-control rules without circumventing authentication, for academic research only~\cite{madanay2025physician,mulgund2020data,dunivin2020gender,gao2012landscape}.
%
Physician names and NPIs were used only at the extraction stage to link reviews to the correct provider, were removed before any analysis, and all reporting and the public release are aggregate and de-identified.
%
The public release comprises aggregate trait distributions and analysis code only, and any request for analysis-level data requires a signed Data Use Agreement and institutional approval.
%
Our results describe patterns in trait scores, not judgments of individual physicians, and archetype labels name positions in trait-score space rather than physician quality.
%
The released artifacts must not be used for individual credentialing, hiring, performance review, or insurance, and any individual-level use would require independent clinical validation; physicians may contact the corresponding author to review or remove specific attributions.
%
These safeguards matter because patients increasingly turn to large language models to research and compare physicians~\cite{littrell2025patients,tebra2026choose,landi2026chatgpt,ayo2024characterizing,mendel2025laypeople}, and such summaries propagate the patterns documented here by default: consistent gender gaps, specialty-context effects that outweigh individual style, and aggregation artifacts.
%
By making these biases visible and releasing a pipeline that supports third-party audits as the underlying models change, this study supports scrutiny by policymakers, regulators, platform designers, and patients.
%
For future work producing individually attributable scores from professional-reputation data, we recommend IRB review that considers data subjects even when data are public, aggregate rather than per-individual release, explicit prohibitions on credentialing and insurance use, and pre-registration of any labels applied to individuals.
%

%
%

%
Longitudinal studies tracking trait stability over physician careers could clarify how practice experience shapes patient-perceived characteristics~\cite{lerch2024model}.
%
Intervention studies could test whether targeted feedback on specific trait scores causally improves patient satisfaction and clinical outcomes~\cite{hojat2011physicians}.
%
Future work could also examine whether the demographic patterns documented here affect physician career advancement and compensation, using outcome data we did not analyze.
%
The pipeline could extend to other trait-outcome relationships: clinical quality metrics, safety outcomes, and physician well-being~\cite{hojat2011physicians,Doyle2013PatientExperience}.
%

%
Together these results provide a national-scale map of patient-perceived physician traits across the U.S. clinical workforce.
%
Patients increasingly meet that workforce through LLM summaries, and the patterns documented here are what those summaries inherit.
%
The pipeline and evaluation framework released here provide the starting point for psychometric validation, fairness audits, and intervention trials linking traits to clinical outcomes.

\section{Methods}


This section describes data collection, LLM-based trait extraction, consistency-check procedures, and statistical analyses.

\subsection{Data Collection and Preprocessing}

%
%

%
We constructed our analysis dataset from a corpus of patient reviews of physicians aggregated from four major U.S. rating platforms: Healthgrades, Vitals, RateMDs, and Yelp.
%
The corpus covers 587{,}562 physicians and 8{,}693{,}563 reviews as of August 1, 2021~\cite{wang2022recency}.
%
These 587{,}562 physicians correspond to 51.5\% of the 1{,}141{,}176 U.S. providers in the underlying NPPES-linked registry used for our analytic cohort construction~\cite{cms_nppes} (Supplementary Table~S1).
%
We restricted analysis to physicians with between 5 and 100 reviews across all platforms, balancing textual signal against bias toward very high-volume providers.
%
This filter yielded 226{,}999 physicians for final analysis.

%
Patient reviews were aggregated at the physician level and concatenated into per-physician text profiles.
%
We extracted physician metadata (gender, medical specialty, geographic location, and review rating scores) from the original dataset and supplemented missing fields from public provider directories.
%
Reviews underwent minimal preprocessing: original patient language and sentiment were retained, and only personally identifiable information and non-English content were removed.
%
Geographic and specialty classifications were taken from NPPES directory data~\cite{cms_nppes}.

%
Data quality checks included validation of physician identifiers and removal of duplicate entries.
%
The final dataset covers physicians across all major medical specialties and U.S. geographic regions.
%
Data collection followed the same paradigm as prior peer-reviewed work analyzing Healthgrades patient reviews at comparable scale~\cite{madanay2025physician}.
%
Physician profile information, star ratings, and written reviews were collected from the publicly-displayed pages of the four platforms via rate-limited HTTP requests, without authentication, account creation, or circumvention of access controls.
%
No individual patient or reviewer was contacted.
%
IRB approval and the aggregate-only release policy are documented in the Ethics Approval and Data Handling subsection below.

\subsection{Top-Down Annotation Protocol}

%
%
We used a top-down trait annotation strategy that processed the full review corpus for each physician as a single context.
%
This contrasted with bottom-up approaches that analyze individual reviews separately and aggregate results.
%
The design used the long-context capacity of modern large language models to form impressions across multiple patient narratives.

Formally, let $P = \{p_1, p_2, ..., p_n\}$ represent the set of physicians in our dataset, and for each physician $p_i$, let $R_i = \{r_{i,1}, r_{i,2}, ..., r_{i,k_i}\}$ denote their complete set of patient reviews, where $k_i$ is the number of reviews for physician $p_i$. Our top-down annotation protocol can be mathematically expressed as:

\begin{equation}
T_i = f(R_i, \Theta)
\end{equation}

where $T_i = \{t_{i,1}, t_{i,2}, ..., t_{i,m}\}$ represents the extracted trait vector for physician $p_i$ across $m$ traits, $f(\cdot)$ is the LLM-based extraction function, and $\Theta$ represents the model parameters and prompt specifications.

This differs from bottom-up approaches~\cite{madanay2024classification} that would compute:

\begin{equation}
T_i^{bottom-up} = g\left(\bigcup_{j=1}^{k_i} f(r_{i,j}, \Theta)\right)
\end{equation}

where individual reviews are processed separately and subsequently aggregated through function $g(\cdot)$.

%
The top-down architecture followed Chain-of-Thought prompting principles: each physician was presented by name alongside their aggregated reviews and analyzed through trait-extraction prompts requiring strict XML output.
%
Physician names were used during the extraction phase only, for correct source attribution; all downstream reporting uses aggregate, anonymized data.
%
This approach explicitly encouraged the model to reason before scoring rather than producing isolated assessments, engaging deeper interpretable reasoning over the text while yielding transparent justifications that can be audited by human experts~\cite{zheng2023judging,wei2022chain}.
%

%
For each trait $t_{i,j}$, the extraction function produces a structured output $t_{i,j} = \{s_{i,j}, e_{i,j}, c_{i,j}, u_{i,j}\}$ where $s_{i,j}$ is a five-point qualitative score, $e_{i,j}$ contains evidence grounded in review excerpts or paraphrases, $c_{i,j}$ measures consistency across multiple reviews, and $u_{i,j}$ assesses sufficiency of supporting evidence.
%
This structured output enabled the model to synthesize contradictory evidence across reviews, identify consistent patterns through multiple patient perspectives, resolve ambiguities using broader contextual understanding, and form holistic assessments that mimic expert evaluation while supporting scalable trait inference from noisy, real-world review data.

\subsection{LLM-Based Agent Implementation}

%
%
We built two model-agnostic trait extraction agents that support the six candidate LLMs evaluated in this study (GPT-4.1, GPT-4o, Claude 3.7 Sonnet, Gemini-2.5 Pro, Gemini-2.5 Flash, Gemini-2.5 Flash-Lite) via LangChain.
%
LangChain is an open-source orchestration library that provides a unified programming interface to heterogeneous LLM APIs.
%
The system used standardized APIs with rate limiting, exponential backoff, and SQLite-based response caching across providers.
%
Exponential backoff retries failed requests with progressively increasing wait times, absorbing transient rate-limit and timeout errors.
%
SQLite-based caching persists model outputs locally so repeated requests with the same input return identical responses, ensuring reproducibility and avoiding redundant API calls.

%
The \texttt{PhysicianBigFiveExtractor} agent implemented a structured prompting framework positioning the model as an ``expert psychologist'' analyzing the Big Five personality dimensions (Openness, Conscientiousness, Extraversion, Agreeableness, Neuroticism) from aggregated patient reviews.
%
The \texttt{PhysicianSubFiveExtractor} agent employed an ``expert analyst'' framework designed to evaluate patient perceptions along five healthcare-specific subjective dimensions: Interpersonal Qualities \& Communication (IQC), Perceived Clinical Competence (PCC), Sensitivity to Patient Satisfaction (SPS), Sensitivity to Clinical Outcome (SCO), and Social Trust Signals (STS).
%

%
Both agents used identical two-message prompting structures with system-level analytical instructions and physician-specific review content, but with domain-appropriate role positioning and trait definitions optimized for their respective assessment frameworks.
%
The BigFive agent emphasized psychological trait inference with detailed trait definitions and clinical interpretation guidelines; the SubFive agent focused on patient decision-making factors with explicit behavioral examples and differentiation criteria between related dimensions.
%
Full prompt specifications for both agents are provided in \hyperref[sec:bigfive_prompt]{Supplementary Note~1} and \hyperref[sec:subfive_prompt]{Supplementary Note~2}.
%

%
Output formatting across both agents was enforced through strict XML schema validation requiring five components per trait: trait name validation, discrete scores (No Evidence, Low, Low to Moderate, Moderate, Moderate to High, High), evidence with direct quotes or paraphrased examples, consistency ratings across multiple reviews, and sufficiency assessments of evidence quality.
%
The SubFive agent used \texttt{<result>} XML tags for streamlined parsing; the BigFive agent used \texttt{<personality>} tags for psychological framework alignment.
%

%
Quality assurance protocols included temperature control (0.0--0.1 for deterministic outputs), structured prompt templates preventing injection attacks, physician name validation for accurate trait attribution, and iterative prompt refinement with retry mechanisms featuring progressive attempt numbering for extraction reliability.
%
This dual-agent implementation enabled comprehensive Big-Five and patient-perceived assessment across the full 226{,}999 physician cohort while maintaining methodological consistency and interpretability across both psychological and healthcare-specific evaluation domains.

\subsection{LLM-as-a-Judge Methodology}

%
%

We implemented an LLM-as-a-Judge framework to score trait extraction quality across multiple models.
%
The judge is Claude Opus 4.7, used as an independent cross-family evaluator.
%
This model lies outside the six-model candidate panel and so avoids within-family judge-candidate circularity.
%
A within-family robustness check using Gemini-2.5 Pro as the judge is reported in \hyperref[sec:cross_family_judge]{Supplementary Note~5.5}.

%
The judge follows the same three-step assessment process for each physician-trait combination.
%
(1) \textbf{Initial Independent Judgment}: the judge analyzes patient reviews without seeing extraction model outputs and generates its own trait assessment with supporting evidence and reasoning.
%
(2) \textbf{Comparative Model Evaluation}: the judge rates each extraction model across five quality dimensions on a 0-10 integer scale.
%
(3) \textbf{Final Integrated Judgment}: the judge combines its model ratings with its initial analysis to produce consensus ratings and reliability estimates.
%

%
The five quality dimensions are: \textit{Evidence Quality} (how well cited evidence supports the trait conclusion), \textit{Reasoning Clarity} (logical coherence of the extraction rationale), \textit{Trait Understanding} (correctness of psychological construct interpretation), \textit{Evidence Specificity} (precision and detail of behavioral observations), and \textit{Conclusion Accuracy} (alignment between supporting evidence and the final trait score).
%
Each dimension receives an integer rating (0-10).
%
The composite score weights evidence quality 25\%, reasoning clarity 20\%, trait understanding 20\%, conclusion accuracy 20\%, and evidence specificity 15\%.
%
%
The judge model outputs XML containing initial trait assessments, per-model evaluations with quality scores and feedback, and final consensus judgments.
%
Consensus judgments include cross-model agreement (0-1 scale) and a reliability estimate (0-1 scale).
%
Cross-model agreement measures convergence among extraction models for the same trait.
%
The reliability estimate combines evidence consistency, model performance variability, and the judge's confidence in the evaluation.

%
The framework scales to thousands of physician profiles with a fixed rubric and reasoning logs that can be audited.
%
The judge's ratings serve as the reference for model selection and benchmarking, with the human consistency check reported below.

\subsection{Human-As-A-Judge Validation Platform}

%
%

%
The human annotation platform replicated the three-step protocol used by the LLM judge: (1) \textbf{Independent Expert Analysis}, in which raters analyzed each physician's review corpus without exposure to model outputs and generated their own trait assessments with supporting evidence and reasoning; (2) \textbf{Systematic Model Evaluation}, in which raters scored each extraction model's output across the same five quality dimensions on the 0--10 integer scale; and (3) \textbf{Consensus Formation}, in which raters synthesized their independent analysis with the model evaluations to produce final consensus ratings and cross-model agreement scores.
%

%
Human raters assessed extraction quality using the same five-dimensional framework employed by the LLM judge: Evidence Quality evaluated how well cited patient narratives supported the trait conclusion; Reasoning Clarity assessed the logical coherence of the extraction justification; Trait Understanding measured accuracy in interpreting the psychological construct; Evidence Specificity rated the precision and clinical relevance of behavioral observations; and Conclusion Accuracy determined alignment between supporting evidence and the final trait score.
%
Each dimension received an integer rating from 0 to 10, combined using the same weighting as the LLM judge (Evidence Quality 25\%, Reasoning Clarity 20\%, Trait Understanding 20\%, Conclusion Accuracy 20\%, Evidence Specificity 15\%) to produce composite performance scores comparable across human and automated evaluators.
%

%
The annotation platform incorporated several quality-assurance mechanisms: standardized training with calibration exercises on reference physician profiles; inter-rater reliability monitoring through duplicate evaluations of overlapping physician subsets; bias controls through randomized presentation orders and blinded model identities; and detailed annotation guidelines with explicit criteria for each quality dimension and trait type.
%
Raters completed structured assessment forms capturing numerical ratings, qualitative feedback, specific evidence citations, and reasoning justifications that paralleled the structured outputs generated by the LLM judge.
%

%
This human-as-a-judge methodology serves multiple functions in our validation framework: providing an independent human consistency check on trait extraction quality across diverse physician profiles and specialties; quantifying inter-rater reliability among expert human evaluators to benchmark automated consistency; validating the LLM-as-a-Judge protocol through direct human-machine agreement analysis; and providing interpretable assessment criteria that can inform future improvements to automated evaluation approaches.

\subsection{Evaluation and Validation Framework}

%
Model performance was assessed through a four-stage protocol combining automated quality scoring with an independent human consistency check.
%
The evaluation dataset is 1{,}197 physicians randomly sampled to cover a range of specialties, geographic regions, and review volumes.
%
The judge (Claude Opus 4.7, defined in the LLM-as-a-Judge subsection above) first produced its own reference assessment of each physician's reviews, then scored each extraction model's output on the five quality dimensions defined there.
%

%
Quality ratings use a 0-10 integer scale, with rubric specifications for each dimension.
%
The judge combined individual quality scores into an overall performance score and recorded its reasoning for each evaluation.
%
Cross-model agreement was recorded as a per-case score in $[0,1]$ summarising how closely the six candidate LLMs converged on each trait.

Model performance was quantified using multiple metrics. For continuous trait scores, we computed:

\begin{equation}
MAE = \frac{1}{n}\sum_{i=1}^{n}|y_i - \hat{y}_i|
\end{equation}

\begin{equation}
RMSE = \sqrt{\frac{1}{n}\sum_{i=1}^{n}(y_i - \hat{y}_i)^2}
\end{equation}

where $y_i$ represents the judge model's reference scores and $\hat{y}_i$ represents each model's predicted scores. For categorical assessments, accuracy rates were calculated as:

\begin{equation}
Accuracy = \frac{\text{Number of correct predictions}}{\text{Total number of predictions}}
\end{equation}

High and Low agreement rates measured the proportion of cases in which models and the judge agreed on extreme trait classifications (scores $>$ 0.75 for High, scores $<$ 0.25 for Low).

%
Trained psychological assessors independently rated 300 physician profiles randomly drawn from the 1{,}197 in the LLM-as-a-Judge dataset.
%
These physicians are stratified sampled across specialties and trait distributions to make the validation set representative of the broader dataset.
%
Raters were blinded to LLM outputs initially, producing independent trait assessments before rating each extraction model's output on the same rubric.
%
The same MAE and RMSE metrics defined above were used, with human reference scores replacing the judge model's scores as ground truth.
%
The random baseline is augmented with a 200-evaluation disagreement-targeted pool (\hyperref[sec:hard_case_validation]{Supplementary Note~5.4}), drawn from the lowest 1.7\% of the cross-model agreement distribution in the cross-family judge dataset.
%
This disagreement-targeted design tests whether LLM-derived and human-derived assessments remain aligned on the hardest cases, where the LLM panel itself disagreed most.

\subsection{Statistical Analysis and Clustering Methods}

%
%

%
\textbf{Statistics and Reproducibility:} All statistical analyses used Python 3.12 with scipy.stats, scikit-learn, and statsmodels.
%
Two-tailed tests were used throughout, with significance level $\alpha = 0.05$ unless otherwise specified.
%
Sample sizes (n) are reported for each analysis, and actual p-values are reported rather than significance thresholds.

%
\textbf{Descriptive Analysis:} Trait score distributions are summarized as mean $\pm$ standard deviation (SD) for normally distributed variables and median [interquartile range] for non-normal distributions.
%
Normality was assessed using Shapiro-Wilk tests ($n<5{,}000$) or Kolmogorov-Smirnov tests ($n\geq 5{,}000$).
%
Kernel density estimation used Gaussian kernels with bandwidth selected by Silverman's rule.
%

%
\textbf{Correlation Analysis:} Pearson correlation coefficients were computed for all 45 pairwise trait combinations on the complete-case sample (physicians with all 10 trait scores present).
%
Multiple comparison correction used Bonferroni adjustment ($\alpha_{adj} = 0.05/45 = 0.0011$).
%
Correlation matrices include 95\% confidence intervals computed via Fisher's z-transformation.
%
Partial correlations controlling for review rating score were computed by residualizing each trait against the per-physician mean review rating score via ordinary least squares, then taking Pearson correlations between the residual pairs.
%

%
\textbf{Group Comparisons:} Gender differences in trait scores were assessed using independent-samples t-tests.
%
Effect sizes are reported as Cohen's d with bootstrap 95\% confidence intervals.
%
Specialty comparisons used one-way ANOVA with $\eta^2$ as the effect size and bootstrap 95\% confidence intervals.

%
\textbf{Clustering Analysis:} K-means clustering was performed on z-standardized Big Five trait scores (n = 92,534) to minimize within-cluster sum of squares:
\begin{equation}
WCSS = \sum_{i=1}^{k}\sum_{x \in C_i} ||x - \mu_i||^2
\end{equation}
where $C_i$ represents cluster $i$ and $\mu_i$ is the centroid.
%
Optimal cluster number (k=4) was determined using the elbow method, which showed a clear inflection point at four clusters (see Supplementary Note~4).
%
Cluster stability was assessed by 200 bootstrap resamples (80\% sample fraction), with Adjusted Rand Index between each resample's clustering and the full-sample assignment.
%
Cluster-shape differences were tested by one-way ANOVA per Big Five trait on ipsatized scores (each physician's grand mean removed across the five traits), with $\eta^2$ as the effect size.
%
Boundary physicians with silhouette scores below 0.1 were identified as cases where cluster assignment was less confident.
%
Kruskal-Wallis tests assessed whether trait distributions differed across the four archetypes.

%
\textbf{Regression Modeling:} The relationship between trait scores and the review rating score was modeled using multiple linear regression:
\begin{equation}
Y = \beta_0 + \sum_{j=1}^{p} \beta_j X_j + \epsilon
\end{equation}
where $Y$ is the review rating score and the 10 $X_j$ are the trait scores (0--1 raw, standardized to z-scores before fitting).
%
The regression sample comprises $n = 83{,}230$ physicians with complete trait and review rating data.
%
Model fit is reported as $R^2$.
%
Standardized $\beta$ confidence intervals are reported as 95\% bootstrap intervals: each replicate resamples physicians with replacement, re-standardizes the 10 trait predictors, and refits ordinary least squares.

%
\textbf{Missing Data:} Missing trait scores were handled by listwise deletion.
%
Effect sizes are interpreted using Cohen's conventions, and confidence intervals were computed by bootstrap resampling with 1{,}000 iterations.

\subsection{Ethics Approval and Data Handling}

%
This research was conducted in accordance with the principles of the Declaration of Helsinki.
%
This study was reviewed by the Johns Hopkins Medicine Institutional Review Board (protocol IRB00464674) and determined to be exempt research under 45 CFR 46.104(d)(4)(i).
%
The exempt determination reflects secondary analysis of publicly available data with no intervention or interaction with participants and no acquisition of identifiable private information.
%
On this basis, the Johns Hopkins Medicine Institutional Review Board determined that informed consent was not required.
%
The human-as-a-judge validation was conducted by trained assessors who were members of the research team, not externally recruited participants.
%
This evaluation activity fell within the same Institutional Review Board exempt determination, and no separate recruitment or informed consent was required.
%
This determination follows the same publicly-available-data basis used in prior peer-reviewed work on online physician reviews~\cite{madanay2025physician,mulgund2020data}.
%
All physician-level results are reported in anonymized, aggregated form; no raw review text is redistributed.
%
Identifiers are handled in two phases: physician names and NPIs are used only during the extraction phase to attribute each review to the correct provider, and are removed before any analysis, so that every figure, table, and released artifact reports only aggregate, de-identified results.
%
Person-level data did not leave the Johns Hopkins School of Medicine, and data were stored on a password-protected research computer accessible only to authorized study-team members.
%



\section*{Data Availability}
%
The datasets generated and analysed during the current study are available from the corresponding author on reasonable request.
%
Raw physician review data may contain privacy-sensitive information.
%
Aggregated trait scores and the statistical analysis data supporting the conclusions are available on request, subject to verification of academic research purpose, a signed data use agreement, and approval from the corresponding author's institutional research administration.

\section*{Code Availability}
%
Custom code for LLM-based trait extraction, statistical analysis, and visualization will be made publicly available upon publication acceptance.
%
The released code is intended for aggregate, nationwide research only and should not be applied to individual physician credentialing, hiring, or performance evaluation without independent clinical validation and appropriate ethical review.
%
Key components include (1) LLM agent implementations for personality trait extraction, (2) data preprocessing and consistency-check pipelines, (3) statistical analysis scripts for subgroup comparisons and clustering, and (4) visualization code for all figures.
%
Prior to publication, code is available from the corresponding author on reasonable request for academic purposes.

\section*{Acknowledgements}
This study was supported by the Hopkins Business of Health Initiative (HBHI) 2023 Pilot Grant. We thank HBHI for their generous support and the computational resources that enabled this research. The funder had no role in study design, data collection and analysis, the decision to publish, or preparation of the manuscript.

\section*{Author Contributions}
J.L. conceived and led the project, developed the LLM pipeline and code infrastructure, performed data analysis, and wrote the manuscript. R.H., A.W., Z.D., J.W., X.Z., and T.X. contributed to data analysis, image creation, and manuscript writing. R.A. and G.G. supervised the project, participated in discussions, and contributed to manuscript writing and revision. All authors reviewed and approved the final version of the manuscript.

\section*{Competing Interests}
The authors declare no competing financial or non-financial interests.



\clearpage
\bibliographystyle{naturemag}
\bibliography{0-MapPhyTrait-npjDM2025}


\clearpage

\setcounter{table}{0}
\setcounter{figure}{0}
\renewcommand{\thetable}{S\arabic{table}}
\renewcommand{\thefigure}{S\arabic{figure}}

\begin{center}
{\LARGE\textbf{Supplementary Information}}\\[0.4em]
{\large Mapping Patient-Perceived Physician Traits from Nationwide Online Reviews with LLMs}
\end{center}
\phantomsection\label{sec:supplementary_information}
\vspace{1em}

\section*{Supplementary Note 1: Big Five Personality Trait Extraction Prompt}
\phantomsection\label{sec:bigfive_prompt}
The following prompt template was used for extracting Big Five personality traits from physician reviews:

\noindent\textbf{System Prompt for Big Five Trait Extraction}\par
\vspace{2pt}\hrule\vspace{4pt}
{\small
\textbf{You are an expert psychologist.}

Based on the provided patient reviews, analyze the Big Five personality traits for the focused physician:
\begin{itemize}
    \item Openness
    \item Conscientiousness  
    \item Extraversion
    \item Agreeableness
    \item Neuroticism
\end{itemize}

\textbf{Output Instructions:}
\begin{itemize}
    \item Keep in mind that the reviews are about a physician, not a patient.
    \item If the reviews contain no evidence for a trait, output "No Evidence" for the score.
    \item When finding evidence for a trait, make sure the evidence is related to the physician, not others.
    \item Output strictly in XML format.
    \item For each trait, you must generate:
    \begin{itemize}
        \item \texttt{<name>}: Must be one of [Openness, Conscientiousness, Extraversion, Agreeableness, Neuroticism].
        \item \texttt{<score>}: Must be selected from [No Evidence, Low, Low to Moderate, Moderate, Moderate to High, High]. If no evidence is found, output "No Evidence". If any evidence is found, output the score that best describes the evidence.
        \item \texttt{<evidence>}: Write 2-3 sentences that combine reasoning with direct quotes or paraphrased examples from the reviews.
        \item \texttt{<consistency>}: Must be selected from [Low, Moderate, High]. High consistency means the trait is consistently mentioned across multiple reviews.
        \item \texttt{<sufficiency>}: Must be selected from [Low, Moderate, High]. High sufficiency means the trait is supported by sufficient evidence from the reviews.
    \end{itemize}
\end{itemize}

\textbf{Formatting:}
\begin{itemize}
    \item Output must be strictly within \texttt{<personality>...</personality>} tags.
    \item Use the following structure exactly for each trait:
\end{itemize}

\textbf{Example:}
\begin{verbatim}
<personality>
    <trait>
        <name>Openness</name>
        <score>Moderate</score>
        <consistency>High</consistency>
        <sufficiency>High</sufficiency>
        <evidence>[CHANGE THIS TO YOUR REASONING. EXAMPLE: While Dr. Smith 
        reviews patient histories thoroughly ("asked a lot of questions"), 
        she reacts rigidly when patients propose ideas ("offended when I 
        requested a blood test"), suggesting moderate openness with limited 
        flexibility.]</evidence>
    </trait>
    ...
</personality>
\end{verbatim}

Strictly follow this format and do not include any extra text outside the XML block.
\par}
\vspace{2pt}\hrule\vspace{8pt}

\noindent\textbf{Human Message Template}\par
\vspace{2pt}\hrule\vspace{4pt}
{\small
The Physician to focus is: \{physician\_name\}

The review related to the physician is: 

\{document\}
\par}
\vspace{2pt}\hrule

\clearpage

\section*{Supplementary Note 2: Healthcare-Specific SubFive Trait Extraction Prompt}
\phantomsection\label{sec:subfive_prompt}
The following prompt template was used for extracting healthcare-specific subjective judgment traits from physician reviews:

\noindent\textbf{System Prompt for SubFive Trait Extraction}\par
\vspace{2pt}\hrule\vspace{4pt}
{\small
\textbf{You are an expert analyst evaluating patient perceptions of physicians based on their online reviews.}

Your task is to analyze the physician along five key subjective dimensions that commonly shape patient decision-making. Use the abbreviations below in the XML \texttt{<name>} tags, but rely on the full trait definitions and examples to guide your judgment.

\textbf{Trait Definitions:}

\textbf{1. IQC - Interpersonal Qualities \& Communication}
\\
This trait reflects how the doctor interacts with patients during the visit, including tone, empathy, clarity, listening, and respectful communication. It focuses on the doctor's behavioral style in the moment.
\begin{itemize}
    \item High IQC: "She explained everything clearly and made me feel comfortable."
    \item Low IQC: "He was cold, barely spoke, and didn't seem to listen."
\end{itemize}

\textbf{2. PCC - Perceived Clinical Competence}
\\
This refers to the patient's subjective impression of the doctor's medical skill, judgment, and professionalism. It reflects whether the doctor comes across as knowledgeable, accurate, and effective.
\begin{itemize}
    \item High PCC: "She diagnosed me quickly and clearly explained the options."
    \item Low PCC: "He misdiagnosed me and seemed unsure."
\end{itemize}

\textbf{3. SPS - Sensitivity to Patient Satisfaction}
\\
This trait measures whether the doctor appears to care about how the patient feels - whether they're respected, satisfied, and emotionally supported. It captures efforts to accommodate patient preferences or emotions.
\begin{itemize}
    \item High SPS: "He asked me if I was comfortable with the plan and if I had any concerns."
    \item Low SPS: "She ignored my discomfort and didn't ask how I felt."
\end{itemize}

\textbf{How it differs from IQC:}
\begin{itemize}
    \item IQC is about \textit{how the doctor communicates or behaves} (e.g., warm, respectful, clear).
    \item SPS is about \textit{whether the doctor is motivated} to ensure patient comfort and satisfaction.
\end{itemize}

\textbf{4. SCO - Sensitivity to Clinical Outcomes}
\\
This trait reflects whether the doctor seems to care about the patient's recovery, treatment success, or long-term health outcome. It includes follow-up contact or comments showing concern for the result.
\begin{itemize}
    \item High SCO: "She called a few days later to check on how I was doing."
    \item Low SCO: "After the visit, I never heard from him again."
\end{itemize}

\textbf{How it differs from PCC:}
\begin{itemize}
    \item PCC is about \textit{how capable} or knowledgeable the doctor appears.
    \item SCO is about \textit{how much the doctor cares} about the result of treatment over time.
\end{itemize}

\textbf{5. STS - Social Trust Signals}
\\
This dimension captures references to the doctor's broader reputation - whether other patients trust them, recommend them, or speak highly of them. It includes social proof or repeated use by families.
\begin{itemize}
    \item High STS: "My whole family goes to her, and she's always amazing."
    \item Low STS: No mention of trust or general recommendation by others.
\end{itemize}

\textbf{Output Instructions:}
\begin{itemize}
    \item Use only the review information to evaluate each trait.
    \item If no evidence exists for a trait, mark it as "No Evidence".
    \item Output must be strictly in XML format, using the following structure for each trait:
\end{itemize}

For each trait, include:
\begin{itemize}
    \item \texttt{<name>}: One of [IQC, PCC, SPS, SCO, STS]
    \item \texttt{<score>}: One of [No Evidence, Low, Low to Moderate, Moderate, Moderate to High, High]
    \item \texttt{<evidence>}: 2-3 sentences combining reasoning with quotes or paraphrased patient language.
    \item \texttt{<consistency>}: [Low, Moderate, High], How consistently the trait appears across multiple reviews.
    \item \texttt{<sufficiency>}: [Low, Moderate, High], Whether enough evidence is present to justify the score.
\end{itemize}

\textbf{Example:}
\begin{verbatim}
<result>
    <trait>
        <name>IQC</name>
        <score>Moderate to High</score>
        <consistency>High</consistency>
        <sufficiency>Moderate</sufficiency>
        <evidence>"She was very kind and explained everything step by step." 
        Several reviews emphasize clear communication and warmth during 
        the visit.</evidence>
    </trait>
    ...
</result>
\end{verbatim}

Do not include any extra text outside the XML block.
\par}
\vspace{2pt}\hrule\vspace{8pt}

\noindent\textbf{Human Message Template}\par
\vspace{2pt}\hrule\vspace{4pt}
{\small
The Physician to focus is: \{physician\_name\}

The review related to the physician is: 

\{document\}
\par}
\vspace{2pt}\hrule

\section*{Supplementary Note 3: Case Study for LLM-Derived Traits}

For each trait-value pair, we sampled one physician with 5--10 reviews and the highest extraction confidence.
%
From that physician's reviews, we kept only those containing language relevant to the target trait.
%
We removed all identifiable information (names, dates, platform details) to preserve privacy, while keeping review indices for reference.
%
Within each selected review, the phrases that served as evidence are shown in bold.
\subsection*{3.1 Openness}

\subsubsection*{Definition} 

Reflects intellectual curiosity, creativity, and receptiveness to new ideas and approaches, traits that may influence diagnostic flexibility and problem-solving 

\subsubsection*{Examples for High / Moderate to High}

\begin{quote}
\#6 \\
She is caring, compassionate, a \textbf{brilliant researcher} and admired by other MDs and hospital staff. She figuratively held my hand during heart valve replacement surgery.
\end{quote}

\begin{quote}
\#7 \\
An excellent and caring physician. I have been her patient for many years and she is consistently, attentive, and responsive. Her interest in me and my overall health is very comforting. She answers all my questions and \textbf{enlists me as a partner in my care}.
\end{quote}

\begin{quote}
\#8 \\
Outstanding. She is very smart, kind, and \textbf{extraordinarily thorough}. Without a doubt she is a \textbf{leader in her field}.
\end{quote}

\subsubsection*{Examples for Low / Moderate to Low}

\begin{quote}
\#0 \\
Her office charged me 500 dollars for a 5 minute visit in which she came in, removed a wart from my finger, and was \textbf{halfway out the door before I could get a word in edgewise} to let her know I had another on my thumb (which I'd already mentioned to the receptionist, the nurse, and probably the security guy in the lobby, even).
\end{quote}

\begin{quote}
\#5 \\
I went to this doctor based on reviews. While I am sure she is competent, her ``bedside manner'' is less than stellar. \textbf{I felt my appointment for an all over skin cancer check was rushed (which it was), like she could've cared less}, and then taking a couple of spots off which was just done in a perfunctory manner. When I told my sister I went for this skin cancer check, she asked me if I was checked between my toes and down my legs. \textbf{I wasn't.}
\end{quote}

\begin{quote}
\#6 \\
Absolute billing nightmare. You've been warned. \textbf{There isn't enough room to describe the complete lack of caring from the doctor and her billing staff.} Worst medical experience. EVER!
\end{quote}

\begin{quote}
\#7 \\
After a long wait and finally being able to see a dermatologist for my extreme case of Seborrheic dermatitis (aka dandruff) on my face, I was prescribed dandruff shampoo and lotion which I had tried in an OTC dosage previously with no results. It didn't help, so I went back a second time as requested only to be told \textbf{there was nothing else that could be done.} Really? One treatment option that literally everyone on the planet knows about didn't work and \textbf{there is literally not a single other possibility?} \textbf{The doctor put zero effort into trying to figure it out.}
\end{quote}

\subsubsection*{No Evidence}

\begin{quote}
\#1 \\
This doctor is my primary doctor. I could ask for none better.
\end{quote}

\begin{quote}
\#2 \\
This doctor is excellent!
\end{quote}

\begin{quote}
\#6 \\
This doctor is the best.
\end{quote}

\subsection*{3.2 Conscientiousness}

\subsubsection*{Definition} 

Denotes organization, reliability, and attention to detail, qualities associated with thoroughness in care and adherence to plans. 

\subsubsection*{Examples for High / Moderate to High}

\begin{quote}
\#4 \\
We have both been seeing this doctor for about a year and a half, my wife with some pretty complicated issues, and we could not be more satisfied.  She is \textbf{available, responsive, thorough and above all, attentive.}  We have never felt rushed seeing her.  And her support staff is excellent, particularly Beverly.
\end{quote}

\begin{quote}
\#5 \\
We feel very lucky to have found this doctor.  She is \textbf{sharp, attentive, and truly cares.}  We have been in contact with many medical professionals over the decades and this doctor really stands out.  Excellent staff too!
\end{quote}

\begin{quote}
\#6 \\
Being a patient with a number of medical issues, it is important to me that my doctor really ``\textbf{hear me'' and provide the best treatment available. This doctor does just that!} She truly is a very good physician!
\end{quote}

\begin{quote}
\#7 \\
Considering it was my first visit and that I was a little apprehensive, I feel it was a pleasant experience. I liked the easy access to the Grand Avenue facility/location. The wait time was more than decent. The wait time for the doctor's presence was very short. I felt she \textbf{spent adequate time with me and truly listened to several of my concerns and provided me with informative responses as well as extra advice.}
\end{quote}

\begin{quote}
\#8 \\
This doctor is \textbf{very professional and very caring}. I would recommend her for all my family and friends.
\end{quote}

\subsubsection*{Examples for Low / Moderate to Low}

\begin{quote}
\#1 \\
She seemed \textbf{uncertain most of the time in terms of how to treat my condition. Another gynecologist informed me later that he would have chosen a completely different course of treatment, and what Easterlin recommended probably aggravated my condition. Also, she didn't notice a pretty serious issue I have;} it was discovered by another doctor.
\end{quote}

\subsubsection*{No Evidence}

\begin{quote}
\#0 \\
Condescending, arrogant, argumentative, kept looking at her watch as if I was taking too much of my time.

Horrible neurologist, unhelpful, and just pushing the newest meds.

Avoid at all costs.
\end{quote}

\begin{quote}
\#3 \\
I think this doctor is not only thorough and knowledgeable, she is also caring and compassionate.  She would definitely still be my neurologist if I had not moved so far away.
\end{quote}

\begin{quote}
\#5 \\
This doctor was recommended by my GP and I am so happy with her.  After my first appointment, I found out she was moving to a new location.  I followed her to her new Ewa office even though it meant an additional half hour drive.  She is worth it.
\end{quote}

\subsection*{3.3 Extraversion}

\subsubsection*{Definition} 

Captures sociability, assertiveness, and enthusiasm, potentially shaping a patient’s experience of engagement and interpersonal energy. 

\subsubsection*{Examples for High / Moderate to High}

\begin{quote}
\#0 \\
My Wife and I found this doctor to be very good at explaining my problem as well as taking wonderful care of me. I was up and about very quickly and pain free before I knew it.  He is \textbf{very personable and easy to talk with.}
\end{quote}

\begin{quote}
\#3 \\
This doctor is fantastic. He did an anterior approach when fixing my mother's hip. I'm a registered nurse and I have been \textbf{very impressed with his care for my mother}.
\end{quote}

\begin{quote}
\#4 \\
A great doctor. One of Brownsville's finest surgeons. The care he gave my 90 year old mother was top notch. \textbf{Great bedside manners} and follow up.
\end{quote}

\begin{quote}
\#5 \\
Dr Quesada was very respectfull and informative. He operated on my wife (ACL replacement, and knee repair), and is the only Dr in the US that does ACL replacement with a manual hand drill, which is more gentle/careful. He is \textbf{always available if we have questions}, and my wife's recovery has been fantastic! Excellent doctor. \textbf{Very respectful, over all very happy!}
\end{quote}

\begin{quote}
\#6 \\
This doctor is a great ortho Dr to have had work on me in Las Vegas NV 7 years ago. I had both hips done by him in 2012. Left side done Nov 12th and Right side 3 weeks later on Dec 3rd. I am doing good to the day at age 44 and still working outdoors as a land surveyor of 20 years+ now. The Anterior Full Hip Replacements were the best thing I ever had done by far from this doctor. Very respectful and \textbf{more important was easy going to make it a comfortable experience.}
\end{quote}

\subsubsection*{Examples for Low / Moderate to Low}

\begin{quote}
\#0 \\
this physician is treating my brother at St Mary for stage 4 COPD. This is our first encounter with him and will be our last. We found him to be \textbf{unapproachable, defensive when asking questions} and unprofessional toward his the others. Long story short, he would spend 5 minutes with him, counter any ideas the hospitalists had to improve his care. One of the house doctors recommened muco mist, he did not, but my brother made a complete turn around from his exacerbation and will be going home
\end{quote}

\begin{quote}
\#3 \\
i HAVE SEEN HIM SEVERAL TIMES AT THE HOSPITAL. HE IS THE WORST DOCTOR I HAVE EVER SEEN. \textbf{I HAD TO GO INTO THE HALL TO ASK HIM A QUESTION HE NEVER LOOKED UP ONLY LOOKED INTO HIS PHONE. SPENDS A MINUTE THEN JUST LEAVES.} THE NURSE SAID MANY PATIENTS CONPLAIN. DO YOURSELF A FAVOR SEE A DOCTOR WHO LOOKS YOU IN THE FACE AND NOT HIS PHONE
\end{quote}

\subsubsection*{No Evidence}

\begin{quote}
\#1 \\
This doctor is my primary doctor. I could ask for none better.
\end{quote}

\begin{quote}
\#2 \\
This doctor is excellent!
\end{quote}

\begin{quote}
\#6 \\
This doctor is the best.
\end{quote}

\subsection*{3.4 Agreeableness}

\subsubsection*{Definition} 

Represents compassion, cooperativeness, and empathy, factors strongly linked to respectful, patient-centered communication. 

\subsubsection*{Examples for High / Moderate to High}

\begin{quote}
\#0 \\
I had a wonderful experience with this doctor. She is \textbf{very warm and caring.}
\end{quote}

\begin{quote}
\#2 \\
This doctor is one of the most caring people I have met. \textbf{She is always willing to take the time with you no matter how busy her day is. You can tell she truly loves what she does and she loves helping others. She really is a doctor that cares for her patients.}
\end{quote}

\begin{quote}
\#5 \\
I was delivering at Sky Ridge and this doctor was my doctor during delivery. Our pre-term baby girl was coming out with her nose up and \textbf{she was able to turn her head inside while I was delivering. I only wanted vaginal delivery, no c-section, and I was so thankful she did everything for me to turn baby's face to right position and saved me from c-section surgery. She did a great job and directed me on how to push right.}
\end{quote}

\subsubsection*{Examples for Low / Moderate to Low}

\begin{quote}
\#0 \\
Far and away the worst psychiatrist I have ever seen.  

I have taken 2 Rx's for several years, which are commonly prescribed to treat their respective conditions. I changed insurance and needed a new provider. I sent medical records 4 days before my appointment. He told me a diagnosis I received years ago was inaccurate, because "most providers don't do it right." \textbf{He does not trust medical records from other providers and does not read them. He came across as patriarchal and aloof, uninterested in my experience with either Rx, or the notes from other board certified doctors, collected over several years. He implied that people frequently sell their Rx drugs on the street.}  

The experience was thoroughly awful. I wouldn't recommend this guy to my enemies.
\end{quote}

\begin{quote}
\#1 \\
I have taken 2 Rx's for several years, the dosages are due to trial and error, documented w prior providers, and have had a profoundly positive effect on my life. He tode me most providers 'don't do it right,' he can't trust records, and \textbf{didn't read mine. And b/c the condition wasn't diagnosed by age 9, the diagnosis is false, and the Rx I have been taking to manage it actually hasn't helped me, at all.}  

\textbf{Incredibly unprofessional, diagnosis came via anecdotes and a 45 min conversation.}  

AVOID.
\end{quote}

\begin{quote}
\#2 \\
\textbf{horrible man with arrogant attitude and harsh manners. He made me cry and made me feel so bad.} I'd never recommend such a man to nay one
\end{quote}

\begin{quote}
\#3 \\
Moves around a lot. He's practiced on the east coast, mid central Illinois, Indiana and now in Chicago @ Norwegian American Hospital. We retained an attorney and Private investigator. \textbf{He won't transfer my daughter to her primary care givers. Misled hospital administrators and lied to staff about calling family members. Daughter is comatose after 1 week of his care. Entered hospital, confused and mobile. Prescribed all the wrong meds. Won't treat her for thrombocytopenia.}
\end{quote}

\subsubsection*{No Evidence}

\begin{quote}
\#1 \\
This doctor was very thorough. She spent over an hour with my husband and I reviewing my symptoms and MRI results. While it took several months to get in for a new patient appointment, she was worth the wait.
\end{quote}

\begin{quote}
\#5 \\
This doctor was recommended by my GP and I am so happy with her. After my first appointment, I found out she was moving to a new location. I followed her to her new Ewa office even though it meant an additional half hour drive. She is worth it.
\end{quote}

\begin{quote}
\#6 \\
During my initial visit a few months ago, it was quickly obvious that this doctor had a detailed level of understanding and plenty of pin point experience in treating my rather rare type of headache condition. After spending 4 1/2 years being treated by another Northern VA neurologist who did not have the experience with my condition that this doctor has, I was fortunate to have been able to reach out, locate her, and have her agree to treat me.
\end{quote}

\subsection*{3.5 Emotional Stability (Neuroticism)}

\subsubsection*{Definition} 

Indicates emotional instability and reactivity, higher levels may be perceived as anxiety or reduced confidence in clinical interactions. 

\subsubsection*{Examples for High / Moderate to High}

\begin{quote}
\#0 \\
In the beginning this doctor seemed to be knowledgeable and caring but in reality he's a self important, pompous jerk. He let me know that he believed he knew more about me than my Hematologist, my Gastroenterologist, my Cardiologist and my PCP of 35 years. \textbf{Beware - he enjoys demeaning his patients!}  
\end{quote}

\begin{quote}
\#2 \\
My dad was in the hospital treated from congestive heart failure, \textbf{he refused to give duiretics to him without any compelling reasons so he developed edema and stopped walking ..}
\end{quote}

\begin{quote}
\#5 \\
This doctor \textbf{misdiagnosed me and then proceeded to shame and humiliate me!} Run from him. He obviously can't survive in private practice
\end{quote}

\subsubsection*{Examples for Low / Moderate to Low}

\begin{quote}
\#0 \\
Dr Willard \textbf{shows a genuine interest in the overall wellbeing of my children. She makes the effort to connect with them and takes the time to discuss my concerns.}
\end{quote}

\begin{quote}
\#3 \\
I love Dr. Jenny. She has been my doctor all 20 years of my life and soon I will be kicked from her patient list because I'll be too old for the pediatrician. \textbf{She has always eased my anxieties about doctors offices and needles. She always asked me questions about my life and school and followed up with me the next time I visited}
\end{quote}

\begin{quote}
\#5 \\
Dr Willard is \textbf{very patient and caring. Both my 14 year old 8 year old love and trust her. I Trust her decisions because she will deeply analyze the issues.} She has my full confidence.
\end{quote}

\subsubsection*{No Evidence}

\begin{quote}
\#1 \\
I took my daughter to see this doctor after going to numerous doctors over an 8 month period for headache and pressure (head) issues. He spent over 2 hours with us at our first visit really talking with us and trying to understand what she was experiencing and everything that had transpired. He also conducted many neurological tests such as coordination tests, etc. I like his approach as well...he started her out on several brain healing vitamins/supplements as well as other non-medicinal tools to help her. She was also prescribed headache medicine to take only as needed. Before this visit she had gone through spinal taps/many MRIs, and some pretty toxic medicines, etc yielding little improvement. She is now progressing and improving! He really cares about his patients and works with them (by really listening and asking questions) to offer the best care possible. We couldn't be more pleased! He is also humble and kind...rare these days.
\end{quote}

\begin{quote}
\#2 \\
This doctor is a great diagnostician. He spent a very long time with me, explaining different options re medicine \& how to lessen the severity of migraines. He is very patient, kind \& extremely smart. I've called \& asked questions after my visit. This doctor \& his staff is very responsive \& understanding. If you have headaches, this is the place to be seen.
\end{quote}

\begin{quote}
\#4 \\
Excellent and compassionate doctor.  
Unbeatable for headache, sleep and pain issues.
\end{quote}

\subsection*{3.6 IQC (Interpersonal Qualities \& Communication)}

\subsubsection*{Definition} 

Encompasses the physician’s tone, clarity, listening, and empathy, reflecting the moment-to-moment communication style during patient encounters. 

\subsubsection*{Examples for High / Moderate to High}

\begin{quote}
\#0 \\
I had a wonderful experience with this doctor. She is very warm and caring.
\end{quote}

\begin{quote}
\#2 \\
This doctor is one of the most caring people I have met. \textbf{She is always willing to take the time with you no matter how busy her day is. You can tell she truly loves what she does and she loves helping others. She really is a doctor that cares for her patients.}
\end{quote}

\begin{quote}
\#5 \\
I was delivering at Sky Ridge and this doctor was my doctor during delivery. Our pre-term baby girl was coming out with her nose up and \textbf{she was able to turn her head inside while I was delivering and saved me from c-section surgery. She did a great job and directed me on how to push right.} Really recommend this doctor
\end{quote}

\subsubsection*{Examples for Low / Moderate to Low}

\begin{quote}
\#0 \\
Far and away the worst psychiatrist I have ever seen.  

I have taken 2 Rx's for several years, which are commonly prescribed to treat their respective conditions. I changed insurance and needed a new provider. I sent medical records 4 days before my appointment. \textbf{He does not trust medical records from other providers and does not read them. He told me that the Rx's I take, which have had a profoundly positive effect on my quality of life (at no point did he ask me about their effects), were actually not helping me at all, and doing far more harm than good.} He came across as patriarchal and aloof, uninterested in my experience with either Rx, or the notes from other board certified doctors, collected over several years.  
\end{quote}

\begin{quote}
\#1 \\
I have taken 2 Rx's for several years, the dosages are due to trial and error, documented w prior providers, and have had a profoundly positive effect on my life. He tode me most providers 'don't do it right,' he can't trust records, and \textbf{didn't read mine. And b/c the condition wasn't diagnosed by age 9, the diagnosis is false, and the Rx I have been taking to manage it actually hasn't helped me, at all. Incredibly unprofessional, diagnosis came via anecdotes and a 45 min conversation.}
\end{quote}

\begin{quote}
\#3 \\
Moves around a lot. He's practiced on the east coast, mid central Illinois, Indiana and now in Chicago @ Norwegian American Hospital. We retained an attorney and Private investigator. \textbf{He won't transfer my daughter to her primary care givers. Misled hospital administrators and lied to staff about calling family members. Daughter is comatose after 1 week of his care. Entered hospital, confused and mobile. Prescribed all the wrong meds. Won't treat her for thrombocytopenia.} Scary MD
\end{quote}

\subsubsection*{No Evidence}

\begin{quote}
\#0 \\
Office could use an overhaul, but does it really matter?
\end{quote}

\begin{quote}
\#3 \\
amazing doctor. Best one I have ever been to.
\end{quote}

\begin{quote}
\#5 \\
Extremely thorough. I will be dissapointed if he ever retires! He is extremely intelligent and really cares for his patients.
\end{quote}

\subsection*{3.7 PCC (Perceived Clinical Competence)}

\subsubsection*{Definition} 

Reflects the patient’s subjective impression of the physician’s medical expertise, professionalism, and confidence in decision-making. 

\subsubsection*{Examples for High / Moderate to High}

\begin{quote}
\#1 \\
He is so friendly and \textbf{knowledgable. We are comfortable asking him any questions we have about our baby's health and he always makes sure that we understand his answers to our questions. He is thorough with his examination of our baby and willing to explain everything he is doing so that we understand how he is caring for our baby.}
\end{quote}

\begin{quote}
\#2 \\
Excellent care for both my children! \textbf{He has a personal understanding for children of all ages I recommend this doctor to anyone looking for a hardworking, trustworthy, caring professional as well as getting down to the root of any cause physician} as well as a Human Being. Thank you Dr. Vuelvas and Staff for making our experience humble and trustworthy.
\end{quote}

\begin{quote}
\#4 \\
This doctor is very compassionate and \textbf{knowledgeable and understanding.} I've been taking my grandson to him since he was about 10-11. I would recommend him to any parent or grandparents.
\end{quote}

\subsubsection*{Examples for Low / Moderate to Low}

\begin{quote}
\#1 \\
In 2013 I met with this doctor about a recently discovered brain tumor discovered after passed out and hit my head and face after falling. I was scared but had my list of suggested questions by the Mayo.Clinic website. When I inquired as to the cause and nature of tumors, Dr Brooker said, "You are old (I had just turned 60 the month before.), useless, so what did I expect?". ... \textbf{I expected him to inform me what I wanted to know. I expected fair unbiased treatment or a referral to someone who would. I expected to be listened to and some questions be answered. I expected a professional since I was paying him.}
\end{quote}

\begin{quote}
\#3 \\
This doctor tries hard to seem polite. If you are looking for answers to your questions, do not waste your time. Dr. Brooker has his own agenda. \textbf{He does not listen to the patient, but rather speaks very quickly on topics which are of interest to him. His knowledge consists of nothing more than what a patient can find on any medical website. It was an absolute waste of time and quite frankly an insult to pay for quality medical care only to be told to read the Mayo website?}
\end{quote}

\begin{quote}
\#4 \\
This provider was awful. \textbf{He did not listen, did not read the notes or records, seemed extremely detached and disinterested, and provided no help or insight. ... He very clearly did a bit of a smirk/chuckle combo. This might be because he was feeling awkward, but this did not come across well at all and made it even more difficult for me to access the care for which I came. I would never recommend this provider.}
\end{quote}

\begin{quote}
\#5 \\
Was sent because of brain tumor, \textbf{did not follow up on recommended MRI two years from ER. He did not read ER notes.}
\end{quote}

\begin{quote}
\#6 \\
I have never written a bad review about anything in my life. But after my apportionment with this doctor I felt I needed to share. I couldn't have hated a doctor more. He seemed unsympathetic in spite of using all the right words. \textbf{He didn't listen and didn't review any of my symptoms with me. He just wanted to talk about one aspect. ... REALLY? that is how you practice medicine, with just looking at one thing and not considering everything as a whole? He told me it was stress and all in my head.} ... How dare he write me off.
\end{quote}

\begin{quote}
\#7 \\
Ended up with thrombosis in my brain. It wasn't 'slow flow' Dr. Brooker and \textbf{you should have moved much faster than the 4 weeks from getting the referral until the CT scan showed it and you went into a panic trying to cover your behind. In my opinion you should not be in medicine. You rather obviously don't really care for your job or your patients.}
\end{quote}

\begin{quote}
\#8 \\
This is the 2nd appt I have had with him. He chooses to not listen. Recommend I read web md website, His answer is to try new meds. How many meds do I need to take. I have been dealing with migraines over 25 yrs. ... \textbf{I know what I wrote!! I have severe debilitating migraines 3-4 days a week and have for many years although he claims I don't have chronic migraines! He tells me I have medicine overuse migraines but then follows up with that I need to take additional medication. He treats me like a test rabbit!}
\end{quote}

\subsubsection*{No Evidence}

\begin{quote}
\#0 \\
My husband was sick and the doctors at the hospital would not give me answers. I ask for help from them and recived not returned call. My husband went into kidney failer and no one cared. He was meds that effected the kidneys and no test were done to watch this. I feel we were pushed under the bus so to speak. I have lost faith in the medical community
\end{quote}

\begin{quote}
\#3 \\
A very caring and compassionate doctor who really cares about his patients I don't have one thing bad to say about him I would recommend him to anyone he has an excellent staff as well they are always cheerful and they too care about there patients.
\end{quote}

\begin{quote}
\#5 \\
I would recommend this doctor to anyone, He has been my Dr for 25 plus years, and when I moved away from Indpls 13 years ago to Northern IN, I still kept him as my Dr. I travel almost 2 hrs one way and will continue to do so. He is a wonderful man and is in the for his patients.
\end{quote}

\subsection*{3.8 SPS (Sensitivity to Patient Satisfaction)}

\subsubsection*{Definition} 

Captures whether the physician is attentive to the patient’s preferences, comfort, and emotional well-being, core to the patient experience. 

\subsubsection*{Examples for High / Moderate to High}

\begin{quote}
\#2 \\
This doctor \textbf{always answers my questions, even explaining things in detail to my daughter about her treatment. I feel she really cares about each patient's well being and takes the time to listen to us.} Very helpful and knowledgeable doctor!
\end{quote}

\begin{quote}
\#3 \\
This doctor and the staff at Advanced Specialty Care, P.C., are \textbf{knowledgeable, caring, supportive and friendly. They are wonderful to work with and accommodate the needs of their patients.} We wouldn't go anywhere else.
\end{quote}

\begin{quote}
\#4 \\
This doctor is very pleasant and knowledgeable. \textbf{She really cares about helping to make you feel better.} Her Nurse/P.A., Molly is also excellent, and very nice.
\end{quote}

\begin{quote}
\#5 \\
This doctor is the best doctor I have ever gone too. \textbf{She is informative, attentive, and truly cares.} I wish that she could be my doctor for everything!
\end{quote}

\begin{quote}
\#6 \\
This doctor \textbf{has a great bedside manner and understanding of my health issues. I really appreciated her care and compassion through some difficult health issues.}
\end{quote}

\begin{quote}
\#7 \\
This doctor is down to earth, relatable, very knowledgeable, \textbf{kind, compassionate and an excellent allergy and asthma care practitioner}
\end{quote}

\subsubsection*{Examples for Low / Moderate to Low}

\begin{quote}
\#0 \\
\textbf{He did not establish a good patient/doctor relationship from the beginning.}
\end{quote}

\begin{quote}
\#1 \\
Dr is always running late. He is running late because \textbf{every patient gets a lecture about the meds they take. Then he'll make the sales pitch for the nerve stimulator.}
\end{quote}

\begin{quote}
\#2 \\
\textbf{Said he does not treat pain}
\end{quote}

\begin{quote}
\#3 \\
\textbf{Has told both my wife and I that he does'nt treat pain}
\end{quote}

\begin{quote}
\#4 \\
He is \textbf{terrible would let him treat my dog!} I wouldn't give him 1 star
\end{quote}

\subsubsection*{No Evidence}

\begin{quote}
\#0 \\
Dr. Gillis only works for insurance companies to determine if you are injured or not. He is paid by insurance companies. He does not take private patients.
\end{quote}

\begin{quote}
\#1 \\
Excellent and thorough.
\end{quote}

\begin{quote}
\#2 \\
Dr. Gillis was the only doctor who was able to diagnose my fibromyalgia.
\end{quote}

\subsection*{3.9 SCO (Sensitivity to Clinical Outcome)}

\subsubsection*{Definition} 

Indicates whether the physician expresses concern for recovery, treatment success, and long-term health, often cited in reviews mentioning follow-up or care continuity. 

\subsubsection*{Examples for High / Moderate to High}

\begin{quote}
\#0 \\
This doctor completed my husband's very tough knee replacement in 2010. Due to three previous knee surgeries on that knee, and a previous bone infection, much of the bone above and below the knee had to be removed and replaced.  \textbf{In the past 6 years this knee has given my husband no trouble. In fact, he is able to exercise on a spin cycle 5 days a week, an hour each day with no pain or problems. Dr. Stenger worked a miracle.}
\end{quote}

\begin{quote}
\#1 \\
\textbf{I am five days out of surgery for full knee replacement. No narcotics for pain. He is fabulous}
\end{quote}

\begin{quote}
\#2 \\
\textbf{Everything has been wonderful I'm 2 1/2 weeks post total knee replacement on the left and not using any assist device. My surgical site is granulating well.} He's the best
\end{quote}

\begin{quote}
\#3 \\
i needed a total hip replacement at the ripe old age of 53.. This doctor isn't only a cool guy, he's an amazing surgeon ßhe GLUED my incision shut, no staples. therefore, it's not his fault if i don't wear a bikini next summer. hahaha. \textbf{first surgery, hopefully last. cuz they won't be as effortless as this one!} thank you Dr. Stenger.. Donna
\end{quote}

\begin{quote}
\#6 \\
This doctor has been a godsend. I found him to be patient and understanding. \textbf{He has done both my knee replacements with no issues. ... The hip pain is 4 times the pain as my knees and can no longer walk without a cane ... I highly recommend this doctor and I pray surgeries begin soon.}
\end{quote}

\begin{quote}
\#7 \\
We absolutely love this doctor. \textbf{He did my husband's knee replacement surgery. It was nice meeting with the doctor prior to surgery as he explained the entire process and he let us know that the recovery would take time.}
\end{quote}

\subsubsection*{Examples for Low / Moderate to Low}

\begin{quote}
\#0 \\
\textbf{It's been 16 days and I'm still waiting to hear back from my lab tests which were delivered to your office 14 days ago.} Either hire more help or fire your staff. Unemployment is at 10\% there are plenty of skilled people out there.
\end{quote}

\begin{quote}
\#3 \\
Will be switching Dr.'s have been a patient for years but \textbf{needed some answers and it took them 3 weeks to respond to several phone calls from me.} As a "Customer" I would never put up with this kind of service.
\end{quote}

\begin{quote}
\#5 \\
... After waiting over two hours, a nurse came to tell me the doctor had 3 emergencies. ... I waited another half an hour, and was told that I would have to wait another five months to reschedule. ... \textbf{Additionally, forcing a patient to wait an extra year for an annual Pap smear showed a total lack of concern for my health.}
\end{quote}

\subsubsection*{No Evidence}

\begin{quote}
\#0 \\
Unwilling to treat long term patients that have or need pain meds. Front office staff unconcerned and indifferent. Nurses apathetic.
\end{quote}

\begin{quote}
\#2 \\
Staff are rude but none can compare to "doctor" rand himself. will try to push tons of alternative medicine on you and insult you at the same time. has a dog running around his office
\end{quote}

\begin{quote}
\#3 \\
We've been going to Rand Family Care for quite a few years now and we love them. This doctor is kind, caring and takes a genuine interest in you! Most all of the staff have been with him for years and I think that says something!!!!
\end{quote}

\subsection*{3.10 STS (Social Trust Signals)}

\subsubsection*{Definition} 

Refers to broader reputational indicators, such as being widely recommended, trusted by families, or praised by other patients, serving as social proof of trustworthiness. 

\subsubsection*{Examples for High / Moderate to High}

\begin{quote}
\#3 \\
\textbf{I had always good experiences with Dr. Molai. Now as she has moved to California, I will miss her.}
\end{quote}

\begin{quote}
\#4 \\
We have both been seeing Dr. Molai for about a year and a half, my wife with some pretty complicated issues, and we could not be more satisfied. … \textbf{And her support staff is excellent, particularly Beverly.}
\end{quote}

\begin{quote}
\#5 \\
\textbf{We feel very lucky to have found Dr Molai. ... Dr. Molai really stands out. Excellent staff too!}
\end{quote}

\begin{quote}
\#8 \\
This doctor is very professional and very caring doctor. \textbf{I would recommend her for all my family and friends.}
\end{quote}

\subsubsection*{Examples for Low / Moderate to Low}

\begin{quote}
\#3 \\
In a rush, couldn't wait to leave, standing by the door, \textbf{arrogant attitude, rude, bad doctor!~should have his license revoked~!!}
\end{quote}

\begin{quote}
\#5 \\
He doesn't have time for you. He wasn't my primary care physician, \textbf{nor would I choose him if he was the only one available, and he wanted to change several of my routine meds for chronic problems. ... That was unethical.}
\end{quote}

\begin{quote}
\#8 \\
He had us rush to another doctor with a diagnosis that had to do emergency surgery, and the other doctor thought he was crazy. Thank god we found a great doc and it is not lantoria. \textbf{He is not qualified to be seeing people and should have his license revoked to practice medicine.}
\end{quote}

\subsubsection*{No Evidence}

\begin{quote}
\#0 \\
My first visit to this doctor, who is no longer a student, was frustrating at best. I had to wait 50 min past my appt time. I was told it was due to her running behind. I expected to have blood drawn for a complete blood workup, but it wasn't ordered. I was told by her that my prescription would be sent to my pharmacy, but when I went to pick it up hours later, it had not been received. I called as soon as the office opened the next morning and was told I had to leave a message and I'd get a callback that day. Now it is days later and I am still waiting for the callback and my Rx still hasn't been sent to my pharmacy. Obviously, my care is not a priority. I will be finding another doctor.
\end{quote}

\begin{quote}
\#2 \\
This doctor did not follow through either time of my visit.
\end{quote}

\begin{quote}
\#4 \\
I went for a yearly physical where they ask you to bend and twist. Kimberly asked if I had anything else going on. I told her I was a weight lifter and pulled a muscle so don't laugh at my flexibility. She said everything was good and send me on my way. Three months later I get a bill from Quadmed. Turns out she charged me for an office visit For a pulled muscle. I diagnosed myself and didn't get a prescription for any meds. I will not be back to her.
\end{quote}

\clearpage

\section*{Supplementary Note 4: Clustering Validation and Optimization}
\phantomsection\label{sec:clustering_validation}
%
We selected the number of physician archetypes via the elbow method on K-means.

\begin{figure}[H]
\centering
\begin{subfigure}[b]{0.48\textwidth}
\centering
\includegraphics[width=\textwidth]{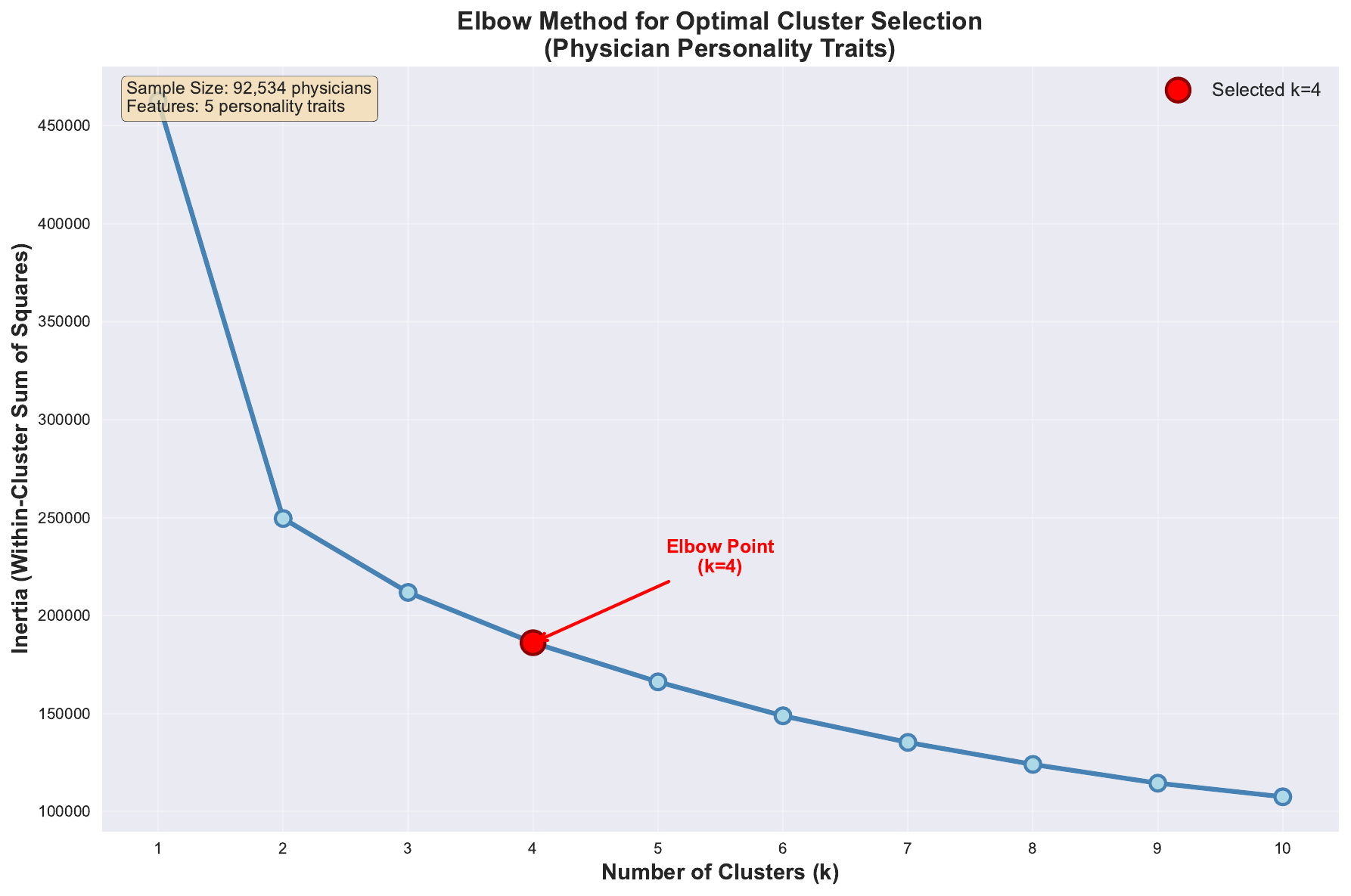}
\caption{Elbow plot for Big Five personality traits}
\label{fig:elbow_bigfive}
\end{subfigure}
\hfill
\begin{subfigure}[b]{0.48\textwidth}
\centering
\includegraphics[width=\textwidth]{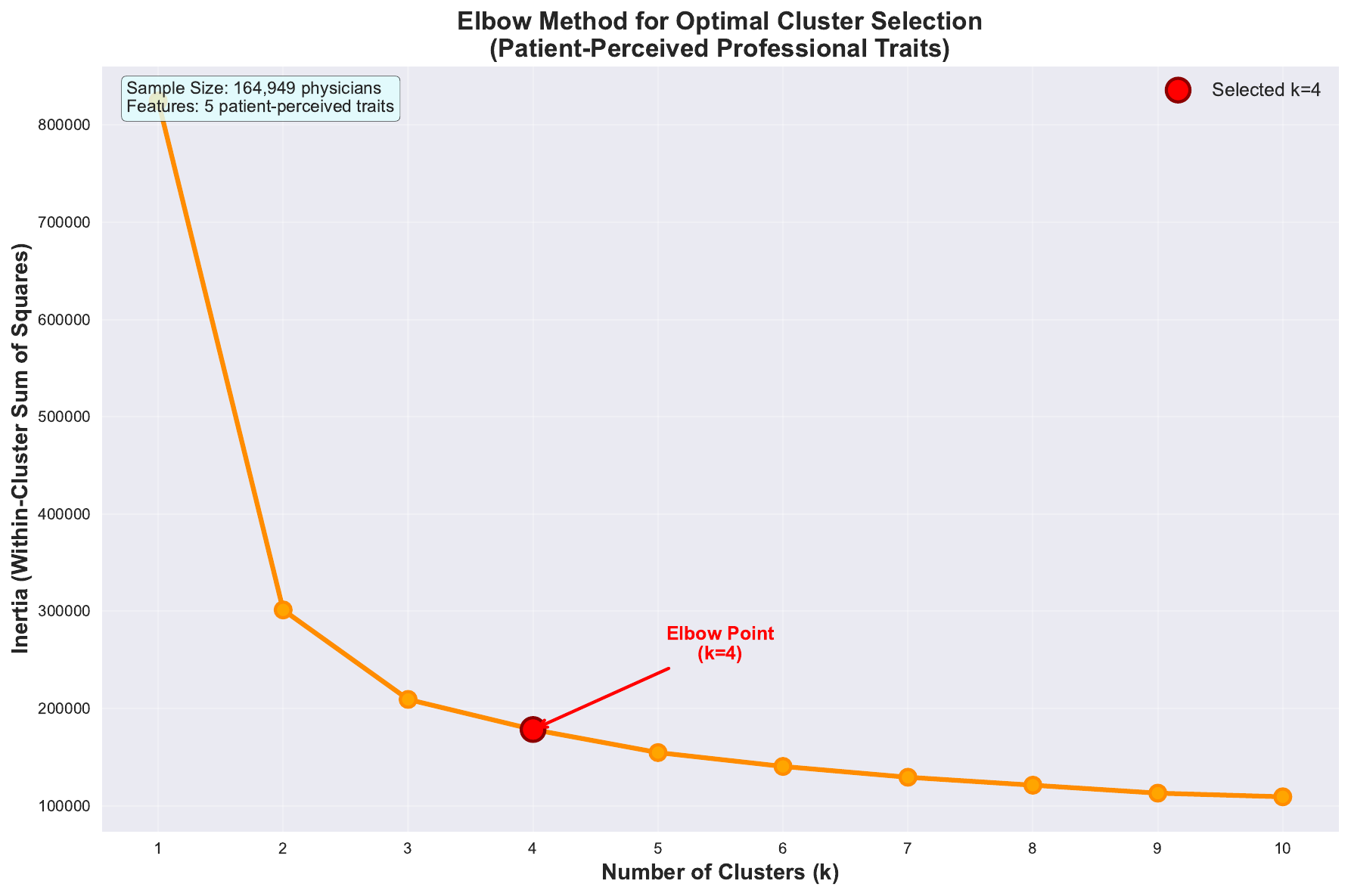}
\caption{Elbow plot for patient-perceived traits}
\label{fig:elbow_subjective}
\end{subfigure}
\caption{Elbow method validation for optimal cluster selection. Both analyses show clear inflection points at k=4 (marked in red), with within-cluster sum of squares (inertia) showing diminishing returns beyond this point. (a) Clustering based on Big Five personality traits (n=92,534 physicians). (b) Validation using patient-perceived professional traits demonstrates consistent optimal k=4 solution. Figure created by the authors using Matplotlib (Python).}
\label{fig:clustering_validation}
\end{figure}

\textbf{Elbow Method:}
%
We computed within-cluster sum of squares (WCSS) for $k = 1$ to $10$ and observed an inflection at $k=4$.
%
WCSS dropped sharply from $k=1$ to $k=4$ with diminishing returns beyond, so four clusters capture the major variance in trait profiles.
%
Big Five and patient-perceived traits independently converged on $k=4$ (Figure~\ref{fig:clustering_validation}).

%
%
The four clusters (``Uniform High,'' ``Uniform Low,'' ``Extraversion Peak,'' and ``Extraversion Valley'') give distinct profiles grounded in the trait data.
%
Cluster labels describe score-pattern positions in trait space, not normative judgments of physician quality.

\section*{Supplementary Note 5: Supplementary Materials Added During Revision}
%
%

\subsection*{5.1 Sample Distribution}

Table~\ref{tab:sample_distribution_overall} summarizes the analytic cohort.
The provider registry covers 1{,}141{,}176 individuals.
After applying the 5--100 review filter used throughout the paper, 226{,}999 physicians (19.9\%) enter the trait-extraction sample.
Per-trait coverage within this cohort ranges from 155{,}252 (Emotional Stability) to 226{,}149 (Agreeableness).
Tables~\ref{tab:sample_distribution_gender}--\ref{tab:sample_distribution_continuous} break the analytic cohort down by gender, specialty, practice state, and continuous variables (star rating, review count, neighborhood Area Deprivation Index).

\begin{table}[H]
\centering
\caption{Overall sample size and per-trait coverage. Registry size (NPPES-linked) and analytic cohort (5--100 review filter). Per-trait coverage is the number of physicians with a non-null trait score within the analytic cohort.}
\label{tab:sample_distribution_overall}
\small
\begin{tabular}{lrr}
\toprule
Characteristic & N & \% \\
\midrule
Total physicians in registry & 1,141,176 & n/a \\
After 5--100 review filter (analytic cohort) & 226,999 & 19.9\% \\
\midrule
Openness coverage & 188,282 & 82.9\% \\
Conscientiousness coverage & 226,114 & 99.6\% \\
Extraversion coverage & 168,471 & 74.2\% \\
Agreeableness coverage & 226,149 & 99.6\% \\
Emotional Stability coverage & 155,252 & 68.4\% \\
IQC coverage & 225,149 & 99.2\% \\
PCC coverage & 226,085 & 99.6\% \\
SPS coverage & 226,065 & 99.6\% \\
SCO coverage & 208,938 & 92.0\% \\
STS coverage & 216,629 & 95.4\% \\
\bottomrule
\end{tabular}
\end{table}

\begin{table}[H]
\centering
\caption{Gender distribution within the analytic cohort (N = 226,999).}
\label{tab:sample_distribution_gender}
\small
\begin{tabular}{lrr}
\toprule
Gender & N & \% \\
\midrule
Male & 158,864 & 70.0\% \\
Female & 68,135 & 30.0\% \\
\bottomrule
\end{tabular}
\end{table}

\begin{table}[H]
\centering
\caption{Top 25 specialties within the analytic cohort (N = 226,999).}
\label{tab:sample_distribution_specialty}
\small
\begin{tabular}{lrr}
\toprule
Specialty & N & \% \\
\midrule
Family Medicine Physician & 35,072 & 15.5\% \\
Internal Medicine Physician & 32,435 & 14.3\% \\
Obstetrics \& Gynecology Physician & 15,998 & 7.0\% \\
Orthopaedic Surgery Physician & 12,404 & 5.5\% \\
Specialist & 10,359 & 4.6\% \\
Pediatrics Physician & 8,537 & 3.8\% \\
Surgery Physician & 8,286 & 3.7\% \\
Psychiatry Physician & 8,046 & 3.5\% \\
Ophthalmology Physician & 7,203 & 3.2\% \\
Gastroenterology Physician & 6,351 & 2.8\% \\
Cardiovascular Disease Physician & 6,204 & 2.7\% \\
Neurology Physician & 5,978 & 2.6\% \\
Dermatology Physician & 5,757 & 2.5\% \\
Urology Physician & 5,080 & 2.2\% \\
Otolaryngology Physician & 4,239 & 1.9\% \\
Neurological Surgery Physician & 3,374 & 1.5\% \\
Hematology \& Oncology Physician & 2,683 & 1.2\% \\
Endocrinology, Diabetes \& Metabolism Physician & 2,670 & 1.2\% \\
Emergency Medicine Physician & 2,535 & 1.1\% \\
Rheumatology Physician & 2,525 & 1.1\% \\
Plastic Surgery Physician & 2,414 & 1.1\% \\
Physical Medicine \& Rehabilitation Physician & 2,223 & 1.0\% \\
(unspecified) & 2,119 & 0.9\% \\
Anesthesiology Physician & 2,028 & 0.9\% \\
Pulmonary Disease Physician & 1,456 & 0.6\% \\
Other specialties & 31,023 & 13.7\% \\
\bottomrule
\end{tabular}
\end{table}

\begin{table}[H]
\centering
\caption{Geographic distribution by practice state (top 20) within the analytic cohort (N = 226,999).}
\label{tab:sample_distribution_geography}
\small
\begin{tabular}{lrr}
\toprule
State & N & \% \\
\midrule
CA & 26,360 & 11.6\% \\
Unknown & 20,891 & 9.2\% \\
FL & 20,869 & 9.2\% \\
TX & 18,609 & 8.2\% \\
NY & 17,513 & 7.7\% \\
PA & 9,364 & 4.1\% \\
IL & 8,826 & 3.9\% \\
OH & 8,801 & 3.9\% \\
MD & 8,586 & 3.8\% \\
MI & 7,753 & 3.4\% \\
GA & 6,748 & 3.0\% \\
AZ & 5,439 & 2.4\% \\
VA & 5,401 & 2.4\% \\
NC & 5,291 & 2.3\% \\
MO & 4,368 & 1.9\% \\
IN & 4,244 & 1.9\% \\
CO & 3,954 & 1.7\% \\
WA & 3,798 & 1.7\% \\
AL & 3,284 & 1.4\% \\
LA & 3,265 & 1.4\% \\
Other states & 33,635 & 14.8\% \\
\bottomrule
\end{tabular}
\end{table}

\begin{table}[H]
\centering
\caption{Continuous variables within the analytic cohort (N = 226,999). Star rating is the patient-rated overall satisfaction score (0--5). Area Deprivation Index (ADI) reflects neighborhood-level socioeconomic deprivation (0--100).}
\label{tab:sample_distribution_continuous}
\small
\begin{tabular}{lrrrrrr}
\toprule
Variable & N & Mean & SD & Median & IQR & Range \\
\midrule
Overall Star Rating & 224,397 & 3.88 & 0.89 & 4.05 & [3.33, 4.60] & [1.00, 5.00] \\
Number of Reviews & 226,999 & 15.87 & 15.20 & 10.00 & [7.00, 18.00] & [5.00, 100.00] \\
Area Deprivation Index & 202,097 & 38.70 & 30.55 & 31.00 & [11.00, 65.00] & [0.00, 100.00] \\
\bottomrule
\end{tabular}
\end{table}

\subsection*{5.2 Partial Correlation Analysis}
\phantomsection
\label{sec:partial_correlation}

%
The full vs.\ sentiment-partialled correlation matrix is shown as Figure~3 in the main text.
%
The analysis ($n = 186{,}250$ physician $\times$ trait records merged with review-rating-score data) reduces the mean inter-trait correlation from $r = 0.61$ to $r = 0.25$ after partialling out the per-physician mean review rating score.
%
All 45 inter-trait pairs remain statistically significant ($p < 0.001$, Bonferroni-corrected) after the partial control.
%
Trait--satisfaction associations attenuate by 66--96\% per trait.
%
Social Trust Signals ($0.81 \rightarrow 0.28$) and Conscientiousness ($0.71 \rightarrow 0.21$) retain the most independent signal; Emotional Stability ($0.66 \rightarrow 0.02$), Sensitivity to Clinical Outcome ($0.74 \rightarrow 0.06$), and Agreeableness ($0.78 \rightarrow 0.07$) are almost entirely explained by sentiment.
%
Bootstrap 95\% confidence intervals (1{,}000 resamples) are narrow throughout (e.g., SPS--satisfaction $r = 0.814$, 95\% CI $[0.812, 0.815]$).

\subsection*{5.3 U-shape Pattern: Sentiment-Artifact Check}
\phantomsection
\label{sec:quality_uShape}

We assess whether the U-shape pattern in sufficiency/consistency versus trait score reflects genuine extraction calibration or an artifact of patient-narrative coherence.
To do this, we computed two internal-consistency metrics per extracted physician-trait score: \emph{sufficiency} (review-corpus adequacy) and \emph{consistency} (cross-review convergence).
Across all traits, sufficiency averaged $0.742$ and consistency averaged $0.627$.

\begin{figure}[H]
\centering
\begin{subfigure}[b]{0.92\textwidth}
\centering
\includegraphics[width=\textwidth]{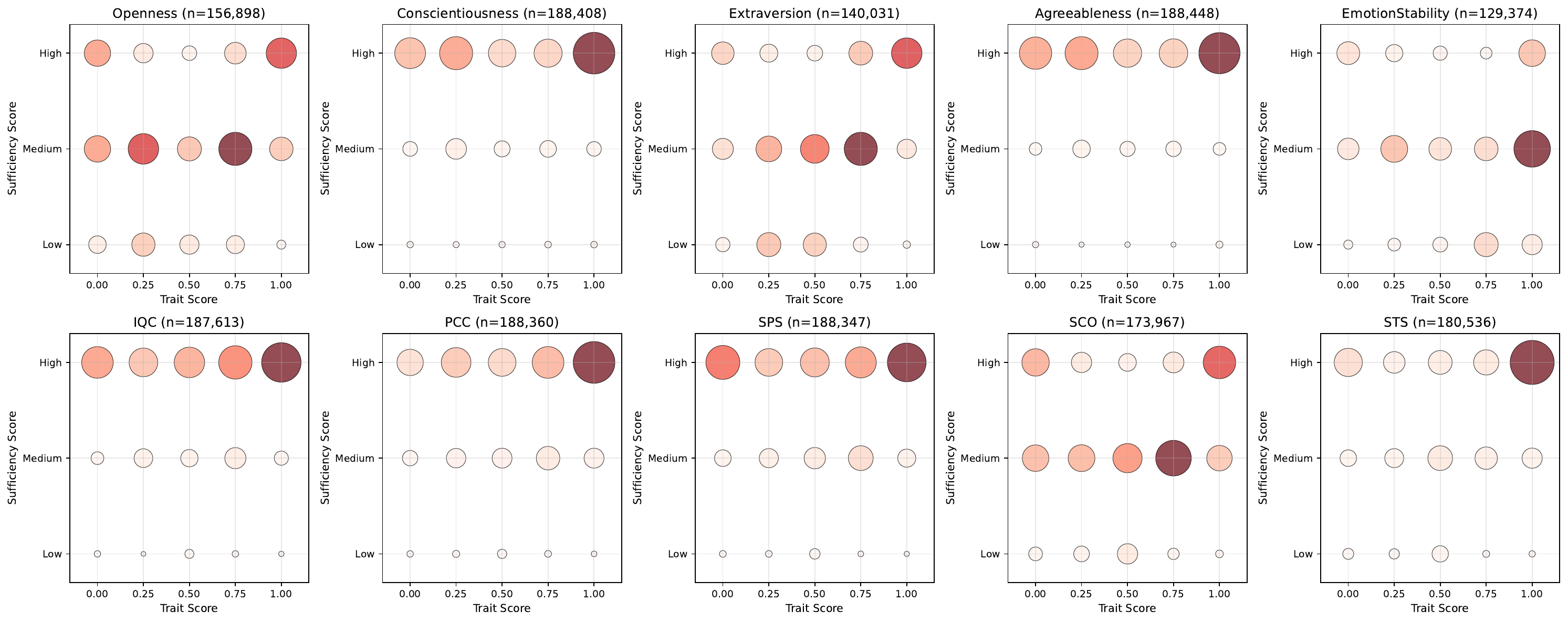}
\caption{Trait score vs.\ sufficiency (raw).}
\label{fig:sufficiency_quality}
\end{subfigure}
\\[0.2cm]
\begin{subfigure}[b]{0.92\textwidth}
\centering
\includegraphics[width=\textwidth]{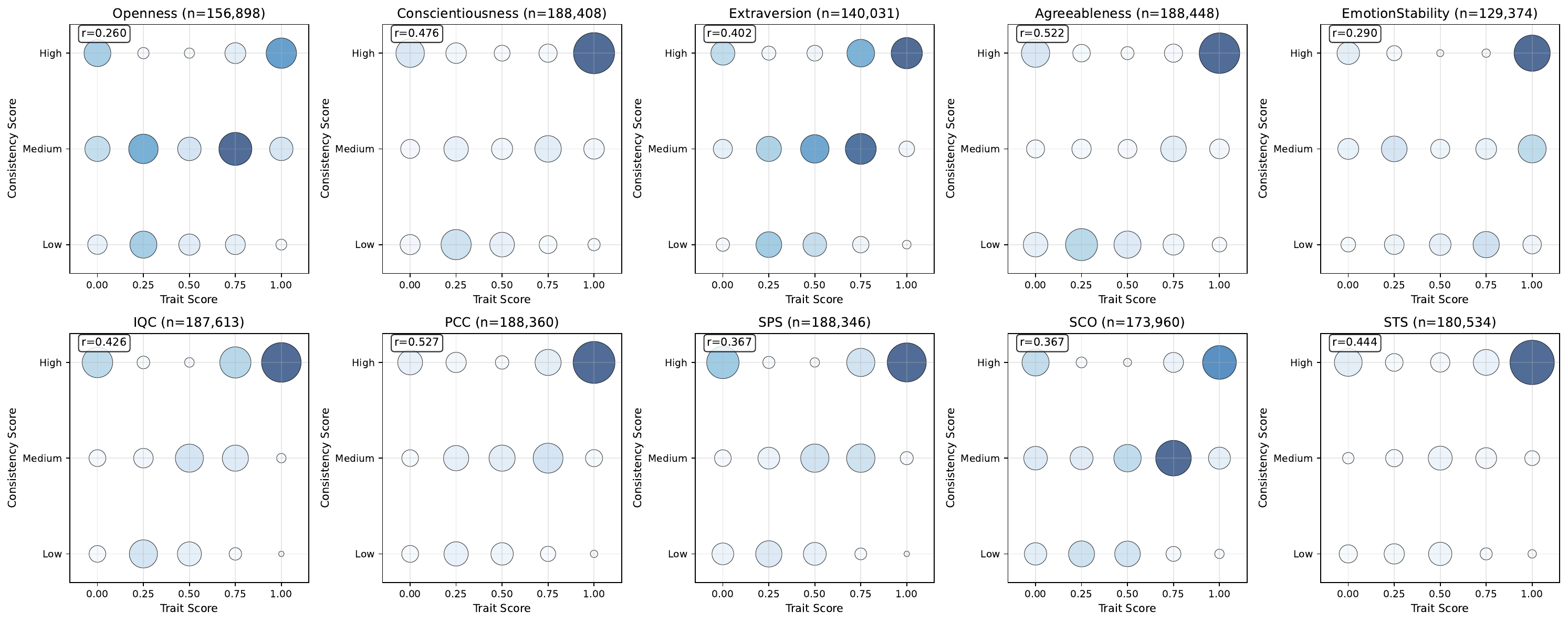}
\caption{Trait score vs.\ consistency (raw).}
\label{fig:consistency_quality}
\end{subfigure}
\caption{\textbf{Raw} trait score vs.\ internal-consistency metrics across all ten traits and 226{,}999 physicians.
(a) Sufficiency.
(b) Consistency.
Data points: trait-by-trait means with SEM; polynomial fits with 95\% confidence bands.
The U-shape is decomposed in Figure~\ref{fig:ushape_residualized}. Figure created by the authors using Matplotlib (Python).}
\label{fig:quality_uShape_raw}
\end{figure}

\textbf{Raw pattern.} Both metrics covary with trait score in a U-shaped pattern (Figure~\ref{fig:quality_uShape_raw}), peaking at extreme trait values ($0$, $1$) and dipping at moderate values ($0.3$--$0.7$).
We originally read this as signal-clarity validation: clearly-expressed traits drive both confident extraction and converging patient descriptions, while moderate scores reflect appropriate uncertainty.

\textbf{Sentiment confound.} An alternative reading is that the raw U-shape reflects the coherence of patient narratives.
Uniformly positive or negative reviews are internally consistent across all trait dimensions, producing high consistency at extreme scores because the underlying patient experience is uniformly valenced, not because the trait was clearly expressed.

\begin{figure}[H]
\centering
\includegraphics[width=0.85\textwidth]{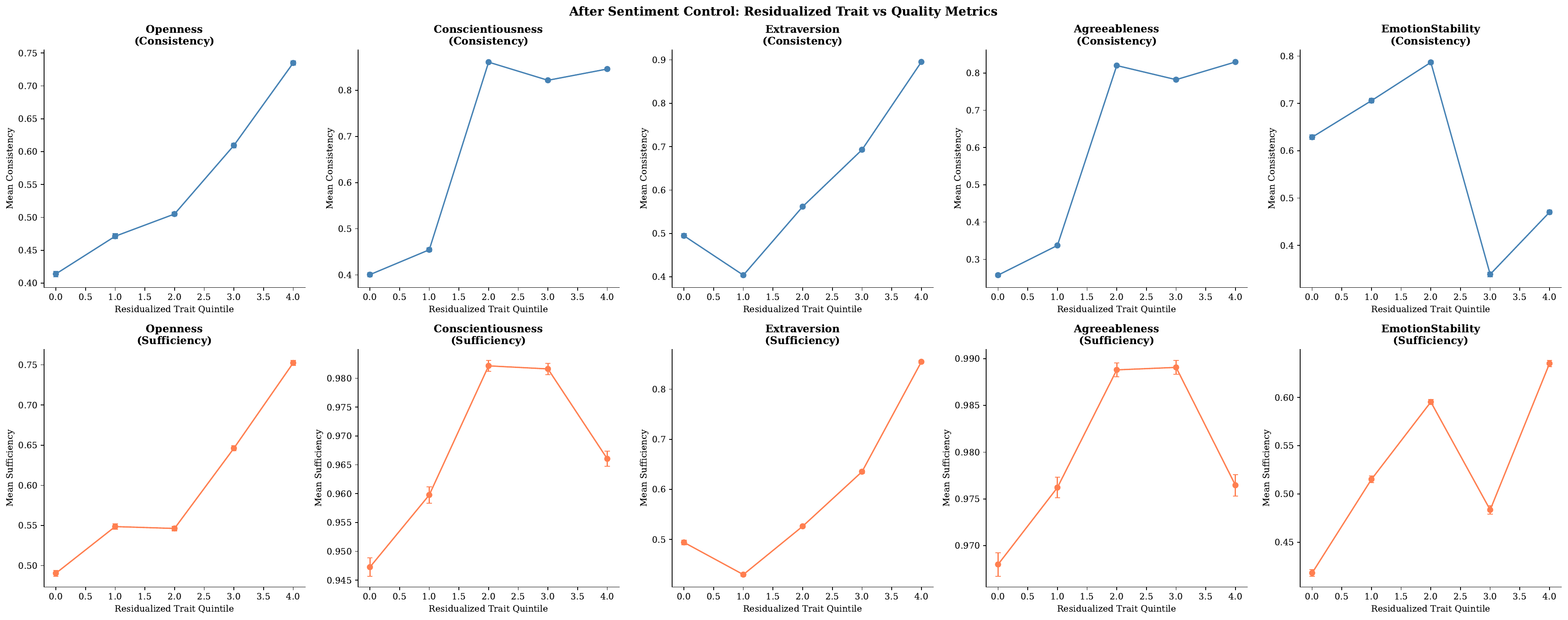}
\caption{\textbf{Sentiment-controlled} trait score vs.\ consistency.
Trait scores are residualised against per-physician mean star rating before re-fitting the quadratic relationship.
The U-shape substantially attenuates or sign-reverses for four of the five Big Five traits and retains $54\%$ of its quadratic coefficient only for Extraversion, indicating that the raw pattern in Figure~\ref{fig:quality_uShape_raw} is mostly a property of patient narrative coherence rather than extraction calibration. Figure created by the authors using Matplotlib (Python).}
\label{fig:ushape_residualized}
\end{figure}

\textbf{Decomposition.} To distinguish these, we residualised trait scores against the per-physician mean review rating score (a sentiment proxy) and re-fit the quadratic relationship between residualised trait score and consistency (Figure~\ref{fig:ushape_residualized}).
Quadratic coefficients before $\rightarrow$ after sentiment control:

\begin{itemize}[leftmargin=1.6em, itemsep=2pt]
 \item \textbf{Extraversion}: $1.89 \rightarrow 0.87$ ($54\%$ retained, sign preserved); partial signal clarity persists.
 \item \textbf{Openness}: $1.87 \rightarrow 0.11$ ($94\%$ attenuation).
 \item \textbf{Conscientiousness}: $2.50 \rightarrow -0.27$ (sign-reversed).
 \item \textbf{Agreeableness}: $2.68 \rightarrow -0.36$ (sign-reversed).
 \item \textbf{Emotional Stability}: $2.67 \rightarrow -0.35$ (sign-reversed).
\end{itemize}

\textbf{Interpretation.} For four of the five Big Five traits, the raw U-shape is largely a structural property of patient review narratives rather than evidence of extraction calibration.
Extraversion is the partial exception: its weaker coupling to overall ratings ($r^2 = 0.022$, vs.\ $r^2 = 0.109$ for Agreeableness) leaves genuine signal-clarity content after sentiment control.
We retain Figure~\ref{fig:quality_uShape_raw} as a description of the raw pattern and Figure~\ref{fig:ushape_residualized} as the sentiment-controlled view.
The partial-correlation analysis (\hyperref[sec:partial_correlation]{Supplementary Note~5.2}) and the cross-family judge evaluation (\hyperref[sec:hard_case_validation]{Supplementary Note~5.4}--\hyperref[sec:cross_family_judge]{5.5}) carry the primary internal-consistency claims in the revised manuscript.

\subsection*{5.4 Disagreement-Targeted Human Validation}
\phantomsection
\label{sec:hard_case_validation}

The random 300-physician sample covers only 0.13\% of the cohort and may be insufficient to detect systematic errors concentrated in specific subpopulations.
To strengthen human validation, we augmented it with a \emph{disagreement-targeted} human-annotation pool drawn from the 11{,}970-case LLM-judge dataset.

\textbf{Pool construction.} For each of the 11{,}970 physician-trait cases, the Opus-judge run records a cross-model agreement score summarizing how closely the six evaluated LLMs agreed on the trait score.
The sample mean is 0.75 (sd 0.17, range 0.30--1.00).
We selected the \textbf{200 cases with the lowest cross-model agreement} (range 0.30--0.55), stratified to 20 cases per trait $\times$ 10 traits, covering 175 unique physicians (16 of whom appear on multiple hard traits).
Five trained annotators each received a balanced assignment of 40 cases (4 per trait), blind to LLM outputs, and re-rated the trait on the same six-level ordinal scale used in the main validation.

\textbf{Two-tier validation.} Combined with the existing 300-physician random sample (300 physicians $\times$ 10 traits = 3{,}000 physician-trait evaluations), this produces a two-tier validation covering \textbf{3{,}200 physician-trait evaluations} spanning the full difficulty spectrum.
The random tier addresses breadth (mean cross-model agreement 0.75); the targeted tier addresses systematic-error detection on the disagreement frontier (agreement $\leq 0.55$).
The 200 targeted evaluations focus on the bottom 1.7\% of the agreement distribution, a region where the 3{,}000-evaluation random baseline would have hit only $\sim$50 cases by chance.

\textbf{Results.} On the 200-case disagreement frontier, the six candidates span MAE 0.207--0.386 against the human annotator score, with Gemini-2.5 Pro leading and GPT-4.1 trailing (main-text Table~2).
For each scorer we restrict to cases where both the scorer and the human annotator assigned a numeric score, excluding ``No Evidence'' labels on either side.
The reported MAE/RMSE/High\%/Low\% metrics match the main-text Table~2, with the human annotator score replacing the cross-family judge as ground truth in this regime.

\textbf{Interpretation.} The hard-case panel produces a regime-specific ordering shift relative to the random panel.
Gemini-2.5 Pro leads on hard-case (MAE = 0.207) and GPT-4.1 trails (MAE = 0.386), reversing the random-panel ordering where GPT-4.1 led (MAE = 0.149) and Gemini-2.5 Pro trailed (MAE = 0.185).
The divergence is informative rather than disqualifying: random and hard-case samples measure different regimes of the trait-extraction task.
On the LLM-Judge panel covering 98\% of the distribution, Gemini-2.5 Pro again leads (MAE = 0.085), so production-scale extraction with Gemini-2.5 Flash rests on its strong performance in the bulk regime and its cost-efficiency, not on hard-case dominance.

\subsection*{5.5 Within-Family Judge Robustness Check (Gemini-2.5-Pro as Judge)}
\phantomsection
\label{sec:cross_family_judge}

Earlier versions of this work used \textbf{Gemini-2.5-Pro} as the LLM-as-a-judge, a within-family choice in which the judge is itself one of the evaluated candidates and may share systematic biases with them.
The revised main-text evaluation (main-text Table~2) uses \textbf{Claude Opus 4.7} (Anthropic family) as an independent cross-family judge, outside the six-model candidate panel.
For completeness, we report here the same protocol run with Gemini-2.5-Pro as the judge so readers can audit how much the cross-family switch changes the rankings.

The two judges yield highly concordant rankings: overall Spearman $\rho = 0.943$ ($p = 0.005$), Kendall $\tau = 0.867$ ($p = 0.017$), and per-trait mean Spearman $\rho = 0.909$ (range $[0.829, 1.000]$).
Gemini-2.5-Pro is ranked first across all 10 traits under both judges, and Gemini-2.5-Flash-Lite is ranked last in 5 of 10 traits.
The only deviation is a swap between Claude~3.7~Sonnet and GPT-4.1 (ranks~3 and~4); the underlying MAE difference is small and consistent with ordinary judge-to-judge measurement noise rather than systematic family bias.

Table~\ref{tab:cross_family_overall} reports the Gemini-as-judge results under the same metric definitions as the main Table~2.
Table~\ref{tab:cross_family_per_trait} reports the per-trait ranking agreement between the two judges.

\begin{table}[H]
\centering
\caption{LLM-as-a-judge results under \emph{Gemini-2.5-Pro as judge} (within-family).
Columns are defined identically to the main-text Table~2: MAE/RMSE are computed on the 0--1 ordinal score (Low=0, Low-to-Moderate=0.25, Moderate=0.5, Moderate-to-High=0.75, High=1); \emph{High} and \emph{Low} are the proportions of items where the model and the judge agree on the extreme classifications ($>0.75$ for High, $<0.25$ for Low).}
\label{tab:cross_family_overall}
\small
\begin{tabular}{lrrrr}
\toprule
Model & MAE & RMSE & High & Low \\
\midrule
gemini-2.5-pro        & 0.0685 & 0.1446 & 79.98\% & 80.21\% \\
gemini-2.5-flash      & 0.1057 & 0.1885 & 91.84\% & 72.56\% \\
gpt-4.1               & 0.1117 & 0.1855 & 88.21\% & 49.78\% \\
claude-3.7-sonnet     & 0.1151 & 0.1929 & 89.36\% & 57.54\% \\
gpt-4o                & 0.1153 & 0.1911 & 86.94\% & 49.78\% \\
gemini-2.5-flash-lite & 0.1517 & 0.2394 & 88.93\% & 49.72\% \\
\bottomrule
\end{tabular}
\end{table}

\begin{table}[H]
\centering
\caption{Cross-family judge: per-trait Spearman agreement between Gemini-judge and Opus-judge model rankings.
The last two columns indicate whether the Opus judge also places Gemini-2.5-Pro at rank~1 and Gemini-2.5-Flash-Lite at rank~6.}
\label{tab:cross_family_per_trait}
\small
\begin{tabular}{lrrcc}
\toprule
Trait & Spearman $\rho$ & $p$ & Pro is \#1 & Flash-Lite is \#6 \\
\midrule
Conscientiousness & 1.00 & 0.000 & yes & yes \\
Openness & 1.00 & 0.000 & yes & yes \\
IQC & 0.943 & 0.005 & yes & yes \\
STS & 0.943 & 0.005 & yes & no \\
Agreeableness & 0.886 & 0.019 & yes & no \\
PCC & 0.886 & 0.019 & yes & no \\
SCO & 0.886 & 0.019 & yes & no \\
SPS & 0.886 & 0.019 & yes & no \\
Extraversion & 0.829 & 0.042 & yes & yes \\
Emotional Stability & 0.829 & 0.042 & yes & yes \\
\bottomrule
\end{tabular}
\end{table}

\subsection*{5.6 Cluster Stability and Profile Shape}

We probe two aspects of the four-cluster solution: bootstrap stability of the cluster assignment and profile shape after ipsatization.

\begin{figure}[H]
\centering
\includegraphics[width=0.75\textwidth]{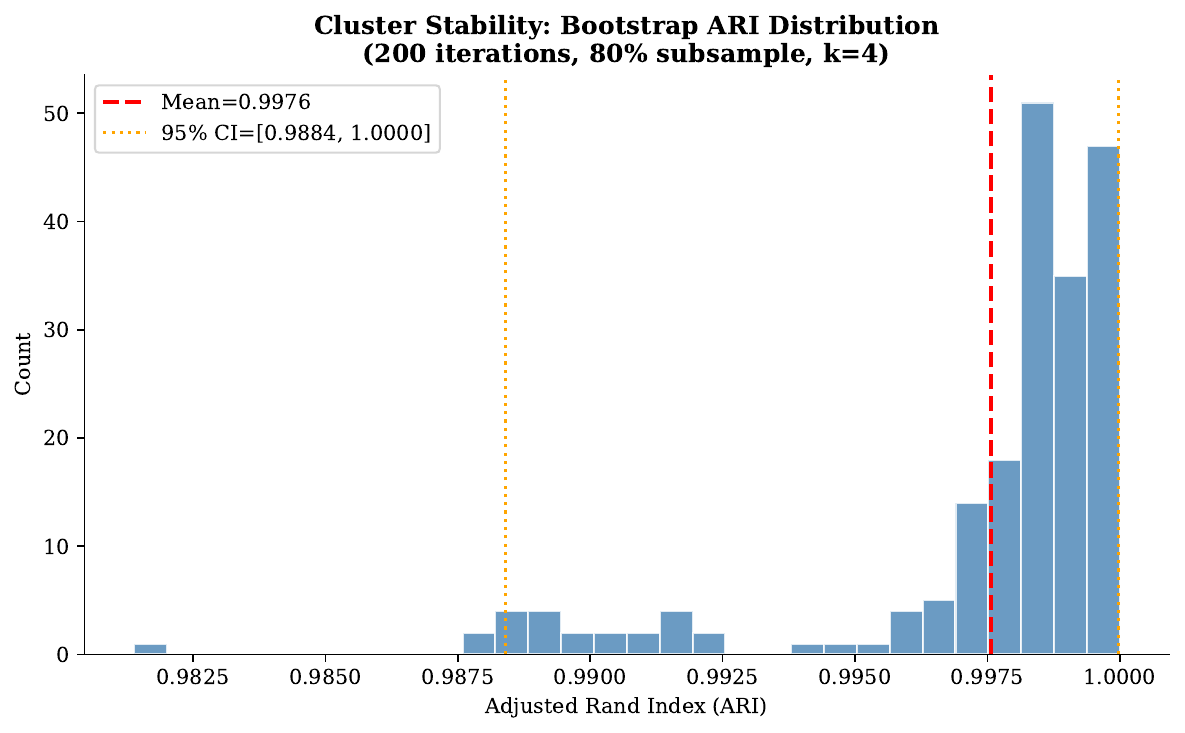}
\caption{Bootstrap distribution of adjusted Rand index between the reference $k{=}4$ clustering and clusterings refit on bootstrap resamples of physicians. Figure created by the authors using Matplotlib (Python).}
\label{fig:cluster_stability}
\end{figure}

\textbf{Bootstrap stability.}
Figure~\ref{fig:cluster_stability} shows the distribution of adjusted Rand index (ARI) between the reference clustering and clusterings obtained on bootstrap resamples of the physician sample.
The distribution is concentrated at high ARI, supporting the stability of the $k{=}4$ partition.

\begin{figure}[H]
\centering
\includegraphics[width=0.95\textwidth]{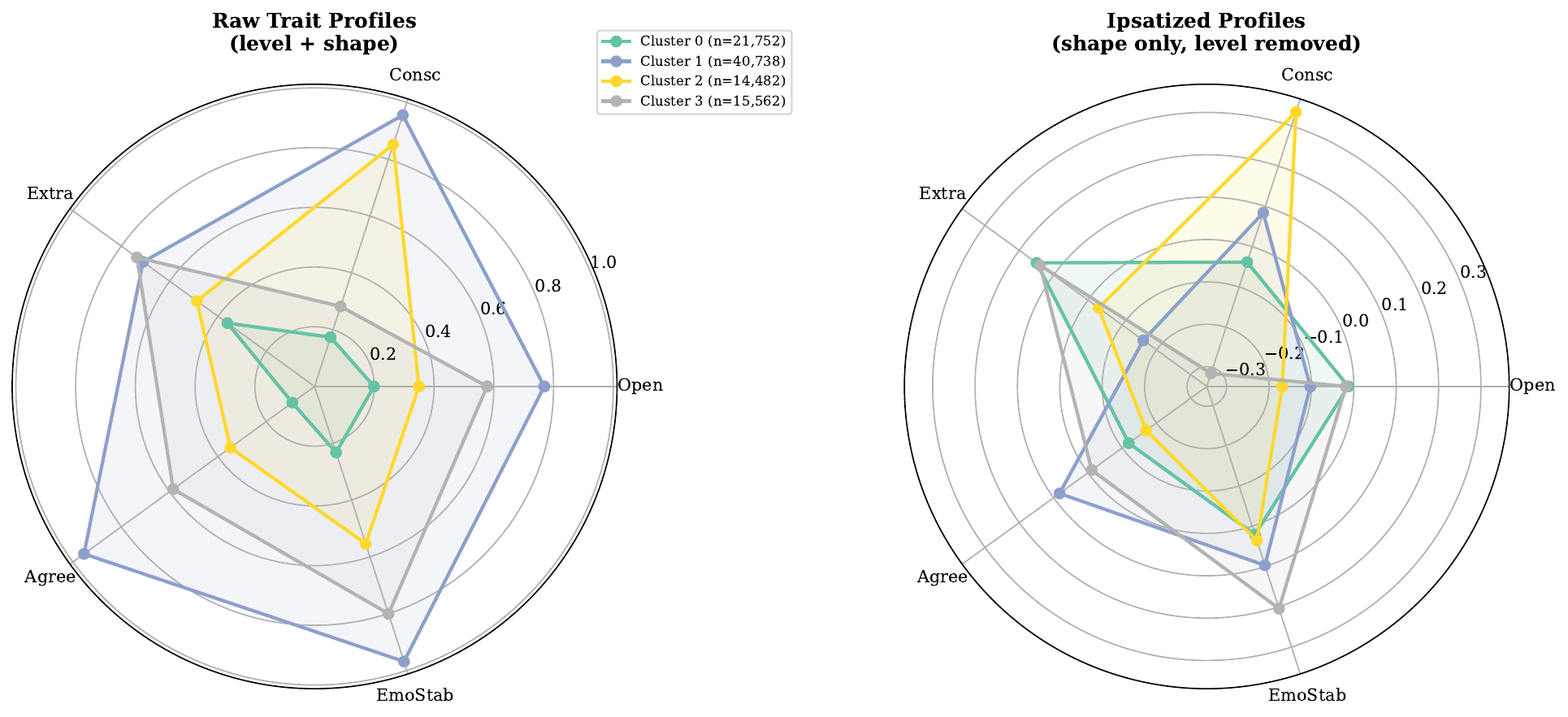}
\caption{Cluster trait profiles before (left) and after (right) within-physician ipsatization.
Pattern-level differences across the four clusters are preserved after removing each physician's overall mean trait elevation. Figure created by the authors using Matplotlib (Python).}
\label{fig:cluster_shape}
\end{figure}

\textbf{Profile shape after ipsatization.}
Figure~\ref{fig:cluster_shape} compares the raw cluster centroids to ipsatized (within-physician centered) profiles.
The shape distinctions between clusters survive ipsatization, indicating that clusters differ in trait \emph{patterning} and not only in overall elevation.

\subsection*{5.7 Bootstrap Confidence Intervals}

All inline 95\% confidence intervals reported throughout the main-text Results (from the \emph{Trait Distributions and Inter-Trait Structure} section through \emph{Cluster Analysis, Latent Profiles}) derive from 1{,}000-resample bootstraps (seed $= 42$; percentile method with 2.5/97.5 percentile cutoffs).
We use paired resampling at the physician level (with replacement) for every estimator that requires aligned multi-column data; one-dimensional resampling is used for univariate statistics.

\textbf{Per-trait means} (\emph{Trait Distributions and Inter-Trait Structure}).
For each of the ten traits we resample the per-physician trait values with replacement and recompute the mean across 1{,}000 replicates; the 2.5/97.5 percentiles give the reported CI.
The unit is one physician $\times$ trait observation; per-trait sample sizes range from 155{,}252 (Emotional Stability) to 226{,}149 (Agreeableness).

\begin{table}[H]
\centering
\caption{Per-trait mean expression with bootstrap 95\% confidence intervals (1{,}000 resamples; physician-level resampling with replacement). Sample size $n$ is the number of physicians with a non-null score for the trait.}
\label{tab:bootstrap_trait_means}
\small
\begin{tabular}{lrcc}
\toprule
Trait & $n$ & Mean & 95\% CI \\
\midrule
Openness & 156,898 & 0.519 & [0.517, 0.520] \\
Conscientiousness & 188,408 & 0.631 & [0.629, 0.633] \\
Extraversion & 140,031 & 0.587 & [0.585, 0.588] \\
Agreeableness & 188,448 & 0.612 & [0.610, 0.614] \\
Emotional Stability & 129,374 & 0.713 & [0.711, 0.715] \\
IQC & 187,613 & 0.631 & [0.630, 0.633] \\
PCC & 188,360 & 0.711 & [0.709, 0.713] \\
SPS & 188,347 & 0.593 & [0.591, 0.594] \\
SCO & 173,967 & 0.578 & [0.576, 0.580] \\
STS & 180,536 & 0.761 & [0.759, 0.762] \\
\bottomrule
\end{tabular}
\end{table}

\textbf{Gender Cohen's $d$} (\emph{Subgroup Analysis, Gender \& Specialty}).
For each trait we resample female and male physician groups independently with replacement, recompute Cohen's $d$ with pooled standard deviation per replicate, and report the percentile CI across 1{,}000 replicates.

\begin{table}[H]
\centering
\caption{Gender Cohen's $d$ (female reference minus male reference) with bootstrap 95\% confidence intervals (1{,}000 resamples; female and male groups resampled independently with replacement; pooled standard deviation per replicate). Negative values indicate that female-rated physicians have lower scores than male-rated physicians.}
\label{tab:bootstrap_gender_cohens_d}
\small
\begin{tabular}{lrrcc}
\toprule
Trait & $n_{\text{F}}$ & $n_{\text{M}}$ & Cohen's $d$ & 95\% CI \\
\midrule
Openness & 52,264 & 104,634 & -0.078 & [-0.088, -0.068] \\
Conscientiousness & 62,012 & 126,396 & -0.070 & [-0.079, -0.060] \\
Extraversion & 44,568 & 95,463 & -0.046 & [-0.057, -0.034] \\
Agreeableness & 62,064 & 126,384 & -0.029 & [-0.038, -0.019] \\
Emotional Stability & 42,902 & 86,472 & -0.136 & [-0.147, -0.124] \\
IQC & 61,877 & 125,736 & -0.064 & [-0.074, -0.053] \\
PCC & 62,021 & 126,339 & -0.177 & [-0.186, -0.167] \\
SPS & 62,151 & 126,196 & -0.061 & [-0.072, -0.051] \\
SCO & 56,245 & 117,722 & -0.164 & [-0.175, -0.154] \\
STS & 59,264 & 121,272 & -0.116 & [-0.127, -0.107] \\
\bottomrule
\end{tabular}
\end{table}

\textbf{Specialty omnibus $\eta^2$} (\emph{Subgroup Analysis, Gender \& Specialty}).
For each trait we restrict to the top-20 specialties for tractability, resample trait--specialty pairs with replacement, recompute one-way ANOVA $\eta^2$ per replicate, and report the percentile CI.

\begin{table}[H]
\centering
\caption{Specialty omnibus $\eta^2$ (one-way ANOVA across the top-20 specialties) with bootstrap 95\% confidence intervals (1{,}000 resamples; trait--specialty pairs resampled with replacement).}
\label{tab:bootstrap_specialty_eta_squared}
\small
\begin{tabular}{lrcc}
\toprule
Trait & $n$ & $\eta^2$ & 95\% CI \\
\midrule
Openness & 121,850 & 0.012 & [0.011, 0.014] \\
Conscientiousness & 146,738 & 0.038 & [0.036, 0.040] \\
Extraversion & 108,686 & 0.024 & [0.023, 0.026] \\
Agreeableness & 146,787 & 0.033 & [0.031, 0.035] \\
Emotional Stability & 100,181 & 0.034 & [0.032, 0.037] \\
IQC & 146,170 & 0.037 & [0.035, 0.039] \\
PCC & 146,697 & 0.053 & [0.050, 0.055] \\
SPS & 146,717 & 0.033 & [0.031, 0.035] \\
SCO & 134,555 & 0.041 & [0.039, 0.043] \\
STS & 140,788 & 0.051 & [0.048, 0.053] \\
\bottomrule
\end{tabular}
\end{table}

\textbf{Trait--satisfaction Pearson $r$} (\emph{Trait-Rating Correlation}).
For each trait we paired-resample (trait score, overall review-rating score) tuples at the physician level ($n = 186{,}250$ merged records), recompute Pearson $r$ per replicate, and report the percentile CI.

\begin{table}[H]
\centering
\caption{Trait--satisfaction Pearson $r$ (per-physician trait score versus mean review rating score) with bootstrap 95\% confidence intervals (1{,}000 resamples; paired physician-level resampling with replacement). All $p < 0.001$.}
\label{tab:bootstrap_trait_satisfaction_r}
\small
\begin{tabular}{lrcc}
\toprule
Trait & $n$ & Pearson $r$ & 95\% CI \\
\midrule
Openness & 154,909 & 0.553 & [0.549, 0.556] \\
Conscientiousness & 186,346 & 0.693 & [0.691, 0.696] \\
Extraversion & 138,167 & 0.412 & [0.407, 0.416] \\
Agreeableness & 186,382 & 0.747 & [0.745, 0.749] \\
Emotional Stability & 127,635 & 0.669 & [0.665, 0.672] \\
IQC & 185,520 & 0.804 & [0.802, 0.806] \\
PCC & 186,265 & 0.767 & [0.765, 0.769] \\
SPS & 186,250 & 0.814 & [0.812, 0.815] \\
SCO & 171,940 & 0.724 & [0.722, 0.726] \\
STS & 178,481 & 0.789 & [0.787, 0.790] \\
\bottomrule
\end{tabular}
\end{table}

\textbf{Standardized OLS regression $\beta$} (\emph{Trait-Rating Correlation}).
We paired-resample physicians with replacement ($n = 83{,}230$ complete-case sample with all ten traits and the overall review-rating score), re-standardize the ten trait scores per replicate, refit OLS, and collect the ten standardized betas; the percentile CI is reported for each predictor.

\begin{table}[H]
\centering
\caption{Standardized OLS regression coefficients ($n = 83,230$) predicting overall review rating score from the ten trait scores, with bootstrap 95\% confidence intervals (1{,}000 resamples; paired physician-level resampling with replacement; trait scores re-standardized per replicate). Rows are ordered by $|\beta|$ from largest to smallest.}
\label{tab:bootstrap_regression_coefs}
\small
\begin{tabular}{lccc}
\toprule
Trait & $\beta$ & 95\% CI & Bootstrap SE \\
\midrule
STS & +0.213 & [0.207, 0.219] & 0.0030 \\
IQC & +0.134 & [0.124, 0.144] & 0.0052 \\
Conscientiousness & +0.132 & [0.128, 0.136] & 0.0022 \\
SPS & +0.122 & [0.111, 0.131] & 0.0053 \\
PCC & +0.104 & [0.098, 0.110] & 0.0031 \\
Agreeableness & +0.064 & [0.058, 0.071] & 0.0032 \\
SCO & +0.044 & [0.039, 0.048] & 0.0024 \\
Openness & +0.043 & [0.039, 0.046] & 0.0018 \\
Extraversion & +0.036 & [0.033, 0.039] & 0.0017 \\
Emotional Stability & +0.015 & [0.010, 0.020] & 0.0023 \\
\bottomrule
\end{tabular}
\end{table}

\end{document}